\documentclass[twoside,11pt]{article}

\usepackage{dnd}
\usepackage{times}
\usepackage{natbib}
\usepackage{pdflscape}
\usepackage{amsmath}
\usepackage{booktabs}
\usepackage[table]{xcolor}
\usepackage{fancyvrb}
\usepackage{listings}
\usepackage{hyperref}

\firstpageno{74}

\dndheading{16(2)}{2025}{74--110}{Ryuichi Sumida, Koji Inoue, Tatsuya Kawahara}{10.5210/dad.2025.203}

\ShortHeadings{Enhancing Long-term RAG Chatbots with Psychological Models}{{Ryuichi Sumida}, Inoue and Kawahara}

\begin{document}

\title{Enhancing Long-term RAG Chatbots with Psychological Models 

of Memory Importance and Forgetting}

\author{\name Ryuichi Sumida \email sumida@sap.ist.i.kyoto-u.ac.jp \\
       \addr Graduate School of Informatics\\
       Kyoto University
       \AND
       \name Koji Inoue \email inoue@sap.ist.i.kyoto-u.ac.jp \\
       \addr Graduate School of Informatics\\
       Kyoto University
       \AND 
       \name Tatsuya Kawahara  \email kawahara@i.kyoto-u.ac.jp\\
       \addr Graduate School of Informatics\\
       Kyoto University}

\editor{David Traum}
\submitted{01/2025}{11/2025}{12/2025}

\maketitle

\begin{abstract}%
This study addresses the issue of what a Retrieval-Augmented Generation (RAG) chatbot should remember and what it should forget, based on findings from psychology. RAG retrieves relevant memories from past interactions to generate responses, and its effectiveness has been demonstrated. As conversations continue, however, the amount of stored memory keeps growing, which not only requires large storage capacity but also risks retaining unnecessary information, potentially deteriorating retrieval performance.
To tackle this problem, we propose LUFY (Long-term Understanding and identiFYing key exchanges), a RAG chatbot that evaluates six distinct memory-related metrics derived from psychological models and real-world data. Instead of simply summing these metrics, it uses learned weights to determine the contribution of each one. By using these weighted scores, the system can prioritize and retain relevant memories while gradually forgetting less important ones during both retrieval and memory management.
To evaluate the effectiveness of LUFY in long-term conversations, we conducted experiments with human participants, who engaged in text-based conversations with three types of chatbots, each using different forgetting mechanisms, for at least two hours. The length of these conversations was more than 4.5 times longer than the longest conversations reported in previous studies. The results showed that prioritizing emotionally engaging memories while forgetting most of the conversation significantly enhanced user satisfaction. Code and Dataset: \url{https://github.com/ryuichi-sumida/LUFY}, Hugginface Dataset:\url{https://huggingface.co/datasets/RuiSumida/LUFY}
\end{abstract}


\section{Introduction}
\label{sec: Introduction}

In human conversations, what we remember remains a mystery. Similarly, in the realm of chatbots, particularly Retrieval-Augmented Generation (RAG) chatbots, the challenge lies not only in retaining long-term memory but also in deciding what to forget.

RAG retrieves related memories from past conversations and uses them to generate a response, and it has been shown to be effective~\citep{xu2022beyond}. Compared to inputting the entire conversation history into a Large Language Model (LLM), retrieving relevant information is not only efficient and effective~\citep{yu2024defense} but also enhances the chatbot's ability to remember and actively utilize the user's past information, improving the level of engagement and contributing to long-term rapport between the chatbot and the user~\citep{Campos2018conversationalmemory}.

As highlighted in previous studies~\citep{choi2023effortless}, however, a challenge arises: as conversations progress, memory constantly increases. This not only demands significant storage space but also involves retaining a lot of unnecessary information, which could lead to degrading retrieval performance. A simple example would be that humans do not remember every single meal they have had in their entire lives; instead remember only their favorites or particularly memorable ones.

This necessitates selective memory in chatbots, much like the human cognitive process, where only the important elements of a conversation are remembered. Studies have shown that humans typically recall only about 10\% of a conversation~\citep{stafford1984conversational}, making it crucial to develop methods that approximate which 10\% of memories are likely to be most useful to retain. This challenge raises the key question: how can chatbots efficiently identify and prioritize these valuable memories, while discarding irrelevant information?

\begin{table}
    \centering
    \begin{tabular}{lccc}
    \hline
    \multicolumn{1}{c}{\textbf{Dataset}}      & \multicolumn{1}{c}{\textbf{Avg. Turns}}  & \multicolumn{1}{c}{\textbf{Avg. Words}} & \multicolumn{1}{c}{\textbf{Conv. Setting}}       \\ \hline
    Daily Dialog~\citep{li2017dailydialog} & 7.9 & 115.3 & Human-Human                  \\
    SODA~\citep{kim2023soda} & 7.6 & 122.4 & Chatbot-Chatbot                  \\
    CareCall~\citep{bae2022keep} & 104.5 & 515.2 & Human-Chatbot \\
    Conversation Chronicles~\citep{jang2023conversation} & 58.5 & 1,054.8 & Chatbot-Chatbot \\
    MSC~\citep{xu2022beyond} & 53.3 & 1,225.9 & Human-Human \\
    \textbf{LUFY-Dataset (Ours)} & \textbf{253.8} & \textbf{5,538.5} & Human-Chatbot\\ \hline
        \end{tabular}     \vspace{1pt}
    \vspace{1pt}
    \caption{Average turns and words per conversation of LUFY-Dataset compared to existing text-to-text dialog datasets. The average length of a conversation in ours is 4.5x that of MSC~\citep{xu2022beyond}, distributed over 4.8x more turns.}
    \label{tab: Statistics compared to existing dialog datasets}
\end{table}

To address this problem, we propose \textbf{LUFY}: \textbf{L}ong-term \textbf{U}nderstanding and identi\textbf{FY}ing key exchanges in one-on-one conversations. Building upon real-world data and psychological insights, LUFY improves existing RAG models by calculating six distinct memory metrics, whereas the conventional simple models like MemoryBank~\citep{zhong2024memorybank} only account for frequency and recency of memory recall. Additionally, LUFY does not merely sum these metrics but uses learned weights to balance their impact. These weighted scores are then used in both the retrieval and forgetting modules, ensuring that the chatbot effectively prioritizes relevant memories while gradually forgetting less important ones.

To empirically validate our proposed model and its effectiveness in enhancing long-term conversational memory, we conducted an experiment involving human participants. In this experiment, participants engaged in text-based conversations with chatbots equipped with different forgetting modules, each for at least two hours. This duration is at least 4.5 times longer than that of the most extensive existing studies of conversational abilities to date, as shown in Table~\ref{tab: Statistics compared to existing dialog datasets}. Our results show that prioritizing arousing memories while discarding most of the conversation content significantly enhances user experience. This enhancement is evaluated by human assessments, sentiment analysis, and GPT-4 evaluations. Furthermore, our method improved the precision of information retrieval by over 17\% compared to the naive RAG system with no forgetting mechanism, highlighting its potential for more accurate and relevant conversational responses.

We developed this comprehensive dataset of human-chatbot conversations along with annotations for the important utterances. Our approach to data collection is designed to uphold rigorous privacy and ethical standards, with consenting participants and robust de-identification procedures where relevant. 

The organization of this paper is structured as follows:
Section~\ref{sec: Psychology of Memorizing Conversations} introduces findings from psychology and explains how to derive metrics from these findings.
Section~\ref{sec: Quantifying Memory Importance} details methods for assessing the importance of user utterances using the metrics introduced in Section~\ref{sec: Psychology of Memorizing Conversations}.
Section~\ref{sec: System Overview} explores the integration of the importance metric and forgetting mechanisms into RAG chatbots.
Section~\ref{sec: User Experiment} presents the user experiment and its results.
Section~\ref{sec: Related Work} reviews relevant prior work in cognitive memory theory and conversational AI.
This paper culminates with conclusions and future works presented in Section~\ref{sec: Conclusion}.

\section{Psychology of Memorizing Conversations}
\label{sec: Psychology of Memorizing Conversations}

MemoryBank~\citep{zhong2024memorybank} introduced the idea of assessing the importance of a memory based on the number of times a memory is used and the elapsed time since the memory was last used. This idea was inspired by the Ebbinghaus Forgetting Curve theory~\citep{ebbinghaus1885ueber}, a theory that the more times information is reviewed or learned, the slower the rate of forgetting. In this model, the importance of a memory is represented as:
\begin{equation}
Importance = e^{-\frac{\Delta t}{S}} ~ .
\label{eq: Forgetting Curve}
\end{equation}
Here, $\Delta t$ represents the time elapsed since the memory was last retrieved, and $S$ is the number of times the memory is retrieved. In MemoryBank, the initial value of $S$ is set to 1 upon its first mention in a conversation. When a memory item is recalled during conversations, $S$ is increased by 1 and $\Delta t$ is reset to 0. Nevertheless, as acknowledged in their study, this represents a preliminary and oversimplified model for memory's importance updating. Most importantly, their work did not include experiments to validate their proposed concept.

Despite the innovative approach taken by MemoryBank, the model's simplicity overlooks critical aspects of how memories are valued and retained. This oversight becomes especially apparent when considering the concept of flashbulb memories, as described by~\citep{brown1977flashbulb}. These memories, such as the news of the 9/11 terrorist attacks, maintain their vividness and strength regardless of recall frequency. The current method of uniformly updating $S$ fails to reflect the intricate nature of memory consolidation and the differential impact of emotionally charged events. 

In the following, we explain pivotal insights from psychology that offer valuable perspectives on conversations, particularly focusing on memory consolidation and retention mechanisms. Additionally, we demonstrate how these findings can be applied to numerically score the importance of an utterance.

First, emotional arousal greatly enhances memory consolidation~\citep{brown1977flashbulb, conway1994formation, mcgaugh2003memory, reisberg2003memory}, a phenomenon well-documented in studies like flashbulb memories~\citep{brown1977flashbulb}. Emotionally charged events, such as dramatic personal experiences like a breakup or receiving college admissions, are more likely to be remembered. We used RoBERTa~\citep{liu2019roberta} finetuned with EMOBANK~\citep{buechel2017emobank}, a 10k text corpus manually annotated with emotion, to measure arousal in user's utterance.

Second, the element of surprise plays a crucial role in memory retention. Events that deviate from expectations are more memorable~\citep{breton2022spatiotemporal}. To quantify this element of surprise, we used the system's utterance as context to evaluate the perplexity of the user's utterance, a measure of how predictable a piece of text is. We employed the GPT2-Large model to calculate the perplexity, using the chatbot's utterance as context.

Third, the concept of Retrieval-Induced Forgetting (RIF) suggests that selectively recalling certain memories strengthens those memories while making related but unmentioned memories more likely to be forgotten~\citep{hirst2012remembering}. Aligning our method with the concept of RIF~\citep{hirst2012remembering}, we retrieved the top 2 memories despite the higher effectiveness observed with the top 5 (see Appendix~\ref{appendix: Top-k Retrieval} for further analysis of top-\textit{k} retrieval). The memory ranked as the most relevant ($R1$) is used for response generation to reinforce its strength, while the second most relevant memory ($R2$) is retrieved but not used for response generation to encourage forgetting, in accordance with RIF principles. 
We note, however, that retrieving the top two memories relative to the user’s current utterance does not guarantee that $R1$ and $R2$ are semantically similar to each other. Cosine similarity is computed relative to the query, not between memories themselves, so $R2$ may not represent a ‘‘competitor’’ memory in the strict psychological sense of retrieval-induced forgetting. Our use of $R2$ therefore approximates the competitive retrieval structure of RIF but does not assume semantic closeness between $R1$ and $R2$, and future work could refine this by identifying memories most similar to $R1$ directly.

Fourth, motivated by prior work suggesting that approximately 70\% of conversational time is devoted to discussing contemporary events~\citep{dunbar1997human}, we incorporated recency as a key factor in our retrieval strategy (Section~\ref{subsubsec: Retrieval Method}). To evaluate this assumption, we conducted an empirical analysis of the dataset collected from our human experiment (described in Chapter~\ref{sec: User Experiment}). Our findings indicate that only 22.9\% of user utterances reference contemporary events, substantially lower than the literature estimate. Further methodological details and results are provided in Appendix~\ref{appendix:contemporary}. Nevertheless, this proportion still constitutes a meaningful share of conversational content, justifying our inclusion of recency in the system design.

Lastly, the capacity for immediate recall in social interactions is surprisingly limited; studies show that individuals typically remember only about 10\% of the content from a conversation~\citep{stafford1984conversational}. While our model adopts this 10\% capacity limit as a starting point, we do not claim it to be a psychologically essential threshold. Rather, it serves as a first approximation that allows us to explore the consequences of limited memory, and future work could investigate alternative thresholds or adaptive selection strategies to refine this modeling choice.

While our metrics focus on observable, surface-level utterance properties such as emotional arousal or surprise, we acknowledge that human memory in dialogue often involves more complex dynamics — including unspoken implications~\citep{fatihi2025unspoken}, avoidance strategies~\citep{endler1994assessment}, and relational cues~\citep{walther1995nonverbal}. Modeling such phenomena would require integrating theories of discourse structure, social cognition, or pragmatic inference, which remain beyond the scope of this study.

Finally, we note an important conceptual distinction relevant to how psychological theories are applied in this work. Research on memory—such as emotional arousal, surprise, or flashbulb memory—typically concerns the encoding of real-world events. However, conversational AI systems do not observe events directly; they observe only their linguistic expression. In this study, we therefore treat an utterance pair as a practical proxy for an event. We acknowledge that a single event may unfold across multiple utterances, and likewise, an individual utterance may reference only part of an event. Our goal is not to equate utterances with events, but to model the cues available to the system through language. Thus, the psychological principles we draw upon are applied to expressed events rather than the events themselves. Future work could extend this approach by modeling multi-utterance segments as unified event representations.


\section{Quantifying Memory Importance}
\label{sec: Quantifying Memory Importance}

\subsection{Metrics for Assessing Memory Importance}
Building upon the psychological principles of emotional arousal, surprise, and retrieval-induced forgetting, we propose the following metrics for determining the importance of memories. Specifically, we translate these following insights into quantifiable metrics:

\begin{itemize}
    \item \textbf{Emotional Arousal ($A$)}: Measured using RoBERTa~\citep{liu2019roberta} fine-tuned with EMOBANK ~\citep{buechel2017emobank}, capturing the intensity of emotions in the user's utterance.
    \item \textbf{Surprise Element ($P$)}: Assessed with perplexity using the GPT2-Large model, representing the unpredictability of the utterance.
    \item \textbf{LLM-Estimated Importance ($L$)}: Derived from the language model's estimation of the importance of the user's utterance.
    \item \textbf{Retrieval-Induced Forgetting ($R1, R2$):} We define $R1$ and $R2$ as counters that track how often a memory was retrieved as the most relevant and the second most relevant, respectively. 

\end{itemize}
Although both emotional arousal and surprise metrics are inspired by psychological theories tied to event memorability, our operationalization derives directly from the user’s linguistic expression. In particular, the arousal metric is computed from utterance text via emotion classification, and surprise is based on linguistic predictability (perplexity). As such, these measures reflect how events are expressed in speech, not just the events themselves.

While the above mentioned metrics draw directly from well-established psychological findings, we incorporate an additional metric—LLM-estimated importance—to capture broader contextual relevance and coherence that may not be easily quantifiable through arousal, surprise, or retrieval patterns alone. This score reflects an LLM's judgment about whether a given exchange will be useful or influential in future conversations, simulating a kind of narrative salience or pragmatic relevance. 
This component is inspired by the “Generative Agents” study~\citep{park2023generative}, which demonstrated that LLMs can be used to simulate aspects of human memory and behavior in a plausible and interpretable manner. Specifically, that study used LLMs to assess the salience of events in an agent’s life based on their likely future consequences. We adopt a similar prompting approach (see Figure~\ref{fig: Prompt for LLM-estimated importance}), framing importance in terms of both personal relevance and conversational utility.
While not derived from a specific psychological theory, this LLM-based metric acts as a learned heuristic that approximates human-like memory prioritization, especially in ambiguous cases where emotion or surprise is low but contextual relevance is high. It serves to complement, rather than replace, our psychologically grounded metrics.

\begin{figure}[ht]
\centering
\begingroup
\footnotesize  
\begin{verbatim}
On a scale of 1 to 10, where 1 is purely mundane
(like brushing your teeth or making your bed)
and 10 is extremely important (like a breakup or a college acceptance),
please rate the importance of the following conversation.

Rate it based on whether it will be useful in later conversations or not.
Here is a summary of the past conversations: {key_summary}
Here is the conversation: {content}
Be sure to provide both a score and a brief reason for your rating.

\end{verbatim}
\endgroup
\caption{Prompt for LLM-estimated importance}
\label{fig: Prompt for LLM-estimated importance}
\end{figure}

\begin{figure}[t]
  \centering
  \includegraphics[width=0.7\linewidth]{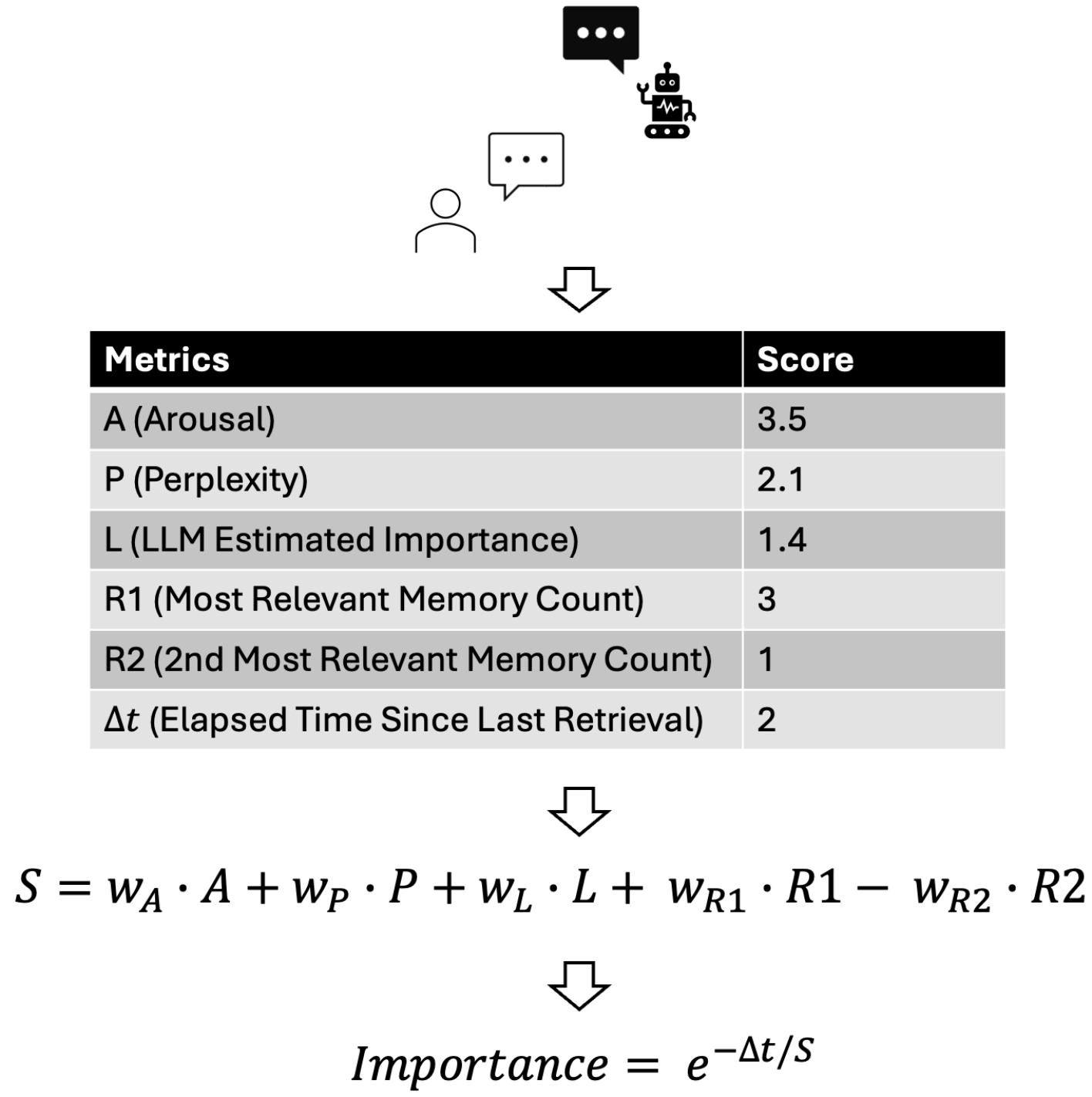}
  \caption{Process of determining the importance of a memory using various psychological metrics. The weights $w_A$, $w_P$, $w_L$, $w_{R1}$, and $w_{R2}$ are assigned based on the relative impact of each metric, as detailed in Section~\ref{sec: Quantifying Memory Importance}.}
  \label{fig: importance_assignment}
\end{figure}

\subsection{Normalization of Metrics} To ensure consistency across the different metrics, we apply a normalization process, ensuring all five metrics—$A$, $P$, $L$, $R1$, and $R2$—are scaled with the same average, minimum, and maximum values. 

Among these metrics, perplexity ($P$) is the only one that can reach notably high values. In cases where perplexity exceeds a threshold of 160, which covers more than 95\% of utterances, the value is capped to avoid extreme outliers. This threshold was selected based on the distribution of perplexity values, but future research could refine this limit to better handle unusually high scores. Using data from the human-RAG chatbot conversations (a set of 200 utterance pairs), we then normalized all metrics to ensure uniform scaling.

\subsection{Definition of Memory Importance}
\label{subsec: Definition of Memory Importance}

As illustrated in Figure~\ref{fig: importance_assignment}, the importance assignment process involves three key steps. 

\textbf{Step 1: Metric Calculation}—For each chatbot-user utterance pair (defined as a memory), the psychological metrics outlined above are calculated. 

\textbf{Step 2: Strength Determination}—Next, the strength $S$ is computed by summing the weighted metrics. 
\begin{equation}
    S = w_A \cdot \textit{A} + w_P \cdot P + w_L \cdot L + w_{R1} \cdot R1 - w_{R2} \cdot R2
\label{eq: S}
\end{equation}

Here, $A$, $P$, and $L$ correspond to metrics for emotional arousal, surprise (perplexity), and LLM-estimated importance, respectively. $R_1$ and $R_2$ are counters indicating how often a memory was retrieved as the most or second-most relevant, respectively. $R_1$ contributes positively to strength, reinforcing the memory, while $R_2$ has a negative effect on strength, in accordance with retrieval-induced forgetting theory. Thus, even though both $R_1$ and $R_2$ stem from retrieval events, they have opposite cognitive effects: $R_1$ supports retention; $R_2$ promotes forgetting.

\textbf{Step 3: Importance Calculation}—Finally, using the calculated strength, the overall importance of a memory is determined. Following the Ebbinghaus forgetting curve formulation introduced in Equation~\eqref{eq: Forgetting Curve}, we compute importance as:

\begin{equation}
\text{Importance} = e^{-\frac{\Delta t}{S}} ~ ,
\label{eq: Importance}
\end{equation}

where $\Delta t$ is the time since the memory was last accessed, and $S$ is the composite strength defined in Equation~\eqref{eq: S}. Although structurally identical to Equation~\eqref{eq: Forgetting Curve}, this formulation reflects a more operational use: $S$ integrates multiple psychologically inspired metrics, going beyond access frequency to capture the nuanced salience of a memory.

\subsection{Weight Estimation}

To quantify the relative impact of each memory metric (Arousal, Perplexity, etc.) on the overall strength, we need to estimate the weights ($w_A$, $w_P$, $w_L$, $w_{R1}$, $w_{R2}$) in Equation \eqref{eq: S}. We achieve this by fitting these parameters to real-world conversational data annotated for importance.

In this study, to fit the three initial parameters ($w_A$, $w_P$, $w_L$), part of the CANDOR corpus~\citep{doi:10.1126/sciadv.adf3197} (a total of 300 utterance pairs, transcribed from audio) was used. While keeping these three parameters fixed, we used human-RAG chatbot conversations (a total of 200 utterance pairs), which were collected prior to the user experiment to fit the latter two parameters ($w_{R1}$, $w_{R2}$). The CANDOR corpus was selected for its diverse set of conversations, while the human-RAG conversations allowed for precise tracking of memory usage, which is not available in the CANDOR corpus.

\begin{table}[ht]
\setlength{\tabcolsep}{3mm}
\centering
\begin{tabular}{l c c}
\hline
Metric & Symbol & Weight \\
\hline
Arousal & \(w_A\) & 2.76 \\
Perplexity & \(w_P\) & -0.28 \\
LLM Estimated Importance & \(w_L\) & 0.44 \\
Most Relevant Memory Count & \(w_{R1}\) & 1.02 \\
2nd Most Relevant Memory Count & \(w_{R2}\) & -0.012 \\
\hline
\end{tabular}     
\vspace{10pt}
\caption{Estimated weights for the Memory-related Psychological Metrics. Arousal has the highest weight relative to other metrics.}
\label{tab: relative importance of the five memory-related metrics}
\end{table}

For both datasets, we tasked annotators with labeling the conversations on a binary scale of 1 and 0, where 10\% of the conversations were labeled as 1 (important) to mimic human behavior~\citep{stafford1984conversational}. Annotators were instructed to label utterances as important if they believed the information contained in the utterance would be used in future conversations. However, we also left some ambiguity, as the notion of what is "important" is not clearly defined, and exploring this ambiguity is part of our study. To assess the reliability of our annotation process, we measured inter-annotator agreement using Fleiss’ kappa~\citep{fleiss1971measuring}. For both the CANDOR corpus and the human-RAG chatbot conversations, three annotators independently labeled each utterance as either important or unimportant. The Fleiss’ kappa score was 0.42 for both datasets, indicating moderate agreement among annotators.

To connect the annotated importance labels (0 or 1) with the strength calculated using the three steps in Section~\ref{subsec: Definition of Memory Importance}, we need to account for the lag time ($\Delta t$). In our setup, the lag time is defined as one time step. This is because the annotation occurs after the conversation (at time $t=2$), and the last time the memory could have been used was during the conversation itself (at time $t=1$). Therefore, we substitute $\Delta t = 1$ into Equation~\eqref{eq: Importance}. This simplifies the equation to:
\begin{equation}
\text{Importance} = e^{-1/S}
\label{eq: Importance_Simplified}
\end{equation}
where S is defined in Equation~\eqref{eq: S} as the weighted sum of the metrics.

With the simplified importance (Equation \eqref{eq: Importance_Simplified}) and the annotated data, we can now estimate the weights. We used the Levenberg–Marquardt algorithm and utilized L2-regularized squared loss as the loss function, with p0 (initial guesses) to be [1, 1, 1], although we did not find the initial guesses to affect the outcome of the learned parameters.

The results of the fitted parameters are given in Table~\ref{tab: relative importance of the five memory-related metrics}. Notably, Arousal (A) exhibits the highest weight among metrics, suggesting that arousal plays a crucial role in determining the memorability and importance of an utterance.

The negative weight for perplexity ($w_P = -0.28$) reflects that lower perplexity correlates with higher importance. This counterintuitive result stems from the nature of the utterances labeled as important. To better understand this, we used an LLM to categorize each utterance as profile-related, episode-related, or others, using the prompt provided in Appendix~\ref{appendix: prompt_profile_episode}. Over half of these utterances (52.3\%) contained profile-related information, such as names, preferences, or other biographical details, which are generally predictable and contextually coherent, resulting in lower perplexity (37.7) compared to the average perplexity of 39.5. Similarly, a significant portion of the utterances (27.3\%) were related to episodes or events, which also exhibited low perplexity (28.7) due to their alignment with prior conversational context. However, the weight of the perplexity is much smaller then those for others. That said, we also recognize limitations in using perplexity as a proxy for psychological surprise. Although perplexity reflects how unexpected a sentence is to a language model, it does not necessarily align with how humans experience surprise or novelty. For example, the sentence ``I saw a car accident right in front of my eyes'' is syntactically conventional and therefore has low perplexity, but the event it describes is rare and emotionally salient. This highlights a key mismatch: perplexity captures linguistic predictability from large-scale training corpora, but not experiential rarity or semantic impact. Thus, while our model finds a weak negative correlation between perplexity and importance in our dataset, this should not be interpreted as a general psychological principle. Future work could explore metrics that better capture event-level novelty or incorporate richer contextual information beyond text.

\section{System Overview}
\label{sec: System Overview}

LUFY is composed of two main components: (1)~the response generation module, and (2)~the forgetting module. The forgetting process is invoked after the conversation when the user has done talking.

\subsection{Response Generation}
\label{subsection: Response Generation}

As depicted in Figure~\ref{fig: response_generation}, the response generation process consists of two main parts: the retrieval part and prompt construction via concatenation of key information (context, summary, and memory). We provide a detailed explanation of the retrieval method, as it is central to the RAG chatbots.

\subsubsection{Retrieval Method}
\label{subsubsec: Retrieval Method}

\paragraph{Memory Storage}
We store three types of information, which together constitute the chatbot's memory:

\begin{figure}[ht!]
\centering
\includegraphics[width=0.8\textwidth]{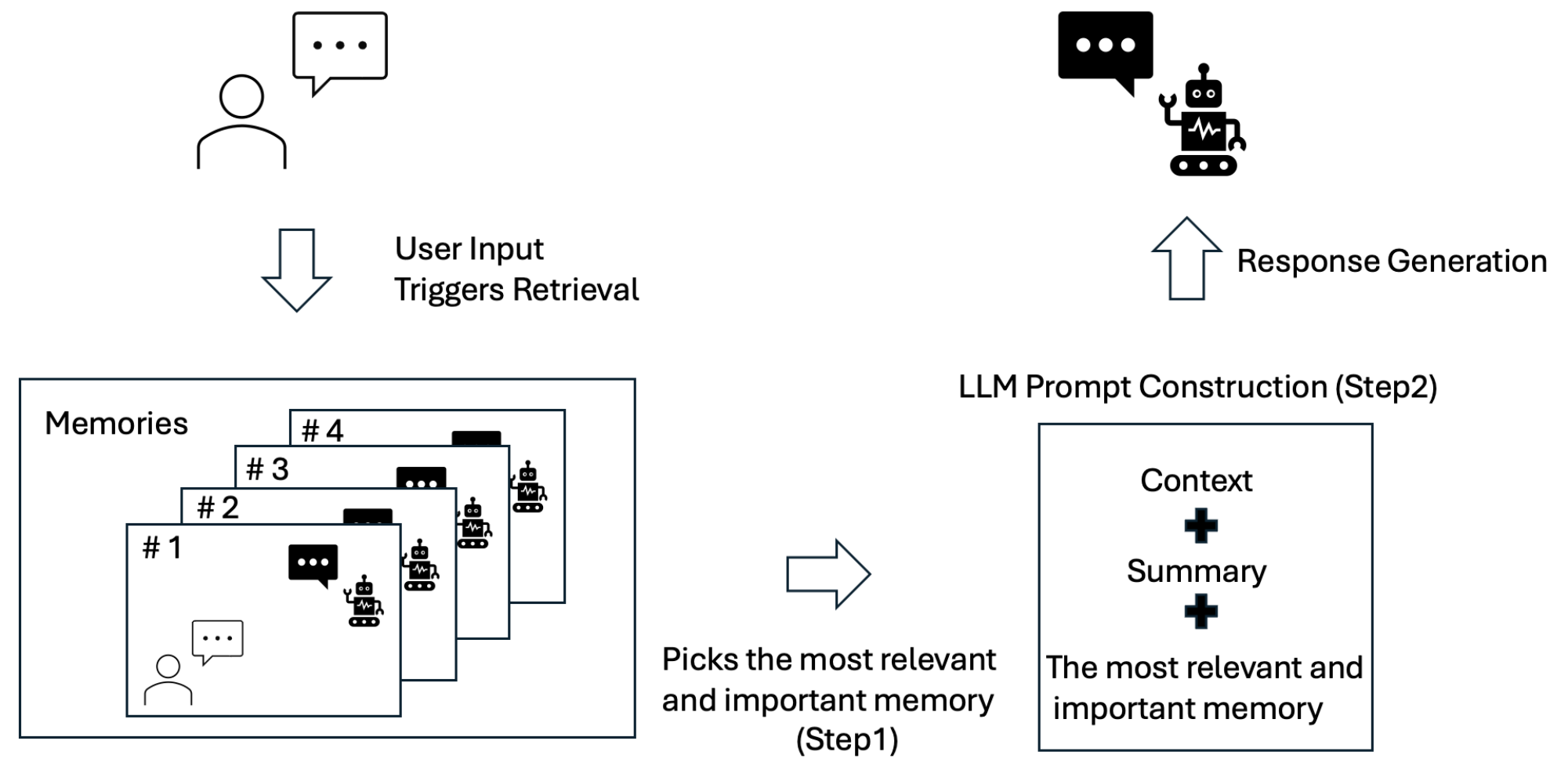}
\caption{Overview of the response generation pipeline. The user input is embedded and compared against stored memory entries based on cosine similarity and an importance score. The top-ranked memory combined with recent utterances and a summary of prior conversations, is passed to the LLM for response generation.}
\label{fig: response_generation}
\end{figure}

\begin{enumerate}
    \item \textbf{Profile information}, consisting of structured facts about the user such as preferences, background traits, or frequently mentioned entities (e.g., ``likes sushi,'' ``has a dog named Luke''). The profile is automatically generated by an LLM based on the accumulated conversation history. It is continuously updated after each session to reflect new salient user details. A ‘session’ refers to a 30-minute human–chatbot interaction (see Section~\ref{sec: User Experiment} for full details).
    
    \item \textbf{Conversation summaries}, which are generated at the end of each session to capture the overall structure and key events of the dialogue.

    \item \textbf{Utterance pairs}, consisting of user--bot exchanges, which preserve granular contextual and episodic details.
\end{enumerate}

We include summaries of the conversations and utterance pairs in memory storage to retain small details about the user that are unlikely to appear in the profile information, such as episodic memory~\citep{tulving2002episodic}, which is made up of past experiences or events they personally remember.

The forgetting mechanism is applied to conversation summaries and utterance pairs, while profile entries are retained and refined over time. This structure enables the system to maintain both stable user traits and specific, temporally grounded memories.

\paragraph{Embedding}
We use a standard LLM embedding (text-embedding-ada-002 from OpenAI). Embeddings of the summaries and utterance pairs are stored.

\paragraph{Retrieval}The foundation of the RAG system is its retrieval method. This process is initiated upon receiving input from a user, to retrieve the memory most relevant to the current conversation. 

There are two key aspects to the retrieval method:
\begin{enumerate}
    \item \textbf{Cosine Similarity Threshold:} During the retrieval stage, we use cosine similarity to assess the relevance between the embeddings of recent conversations and the entries in the memory database. To ensure a high level of contextual relevance, we have set the threshold at 0.8, consistent with commonly used standards in popular RAG system frameworks like LlamaIndex. This choice is further validated by our analysis of Question and Answer pairs collected from the user experiment. The details are available in Appendix~\ref{appendix: Optimal Cosine Similarity Threshold}.

\item \textbf{Final Retrieval Score:} Our proposed method distinguishes itself from previous works by integrating importance into the retrieval process. The final retrieval score is computed using a formula as follows:
\begin{equation}
 \text{score} = \text{Cos. Sim.} + \alpha \cdot \text{Importance}
\end{equation}

Since cosine similarity values are distributed between 0.8 and 0.9 due to our threshold (a range of 0.1), while importance scores range from 0 to 1, we set $\alpha = 0.1$ to approximately normalize their influence in the final score. This choice does not assume that cosine similarity and importance are inherently of equal weight; rather, it heuristically balances the two terms given their different scales. We acknowledge that $\alpha$ is a tunable hyperparameter, and further empirical tuning—e.g., via ablation or grid search—could yield better retrieval performance. This remains an area for future exploration.
\end{enumerate}

\begin{figure}[ht]
\centering
\begingroup
\footnotesize  
\begin{verbatim}
You will be provided with 3 pieces of information:

1. Key Summary: A summary of past conversations .
2. Recent Utterances: The latest exchanges between you and the user.
3. Relevant Memory: The memory most pertinent to the current conversation.

Using these details, you are tasked with generating an effective response. 
Ensure that your reply maintains a casual tone to mimic 
a genuine interaction with a friend. 

Here's the information:

- Key Summary of Past Conversations: {key_summary}
- Recent Utterances: {recent_utterances}
- Memory Relevant to Current Conversation: {related_memory}

\end{verbatim}
\endgroup
\caption{Prompt for generating chatbot responses.}
\label{fig: prompt for generating responses}
\end{figure}

\subsubsection{Information Provided to the LLM for Response Generation}

The LLM is provided with three key pieces of information to generate responses: a summary of past conversations, the context (recent five utterances), and memory relevant to the current context. The prompt given to the LLM to generate a response is shown in Figure~\ref{fig: prompt for generating responses}. 

\begin{figure}
\centering
\includegraphics[width=0.8\textwidth]{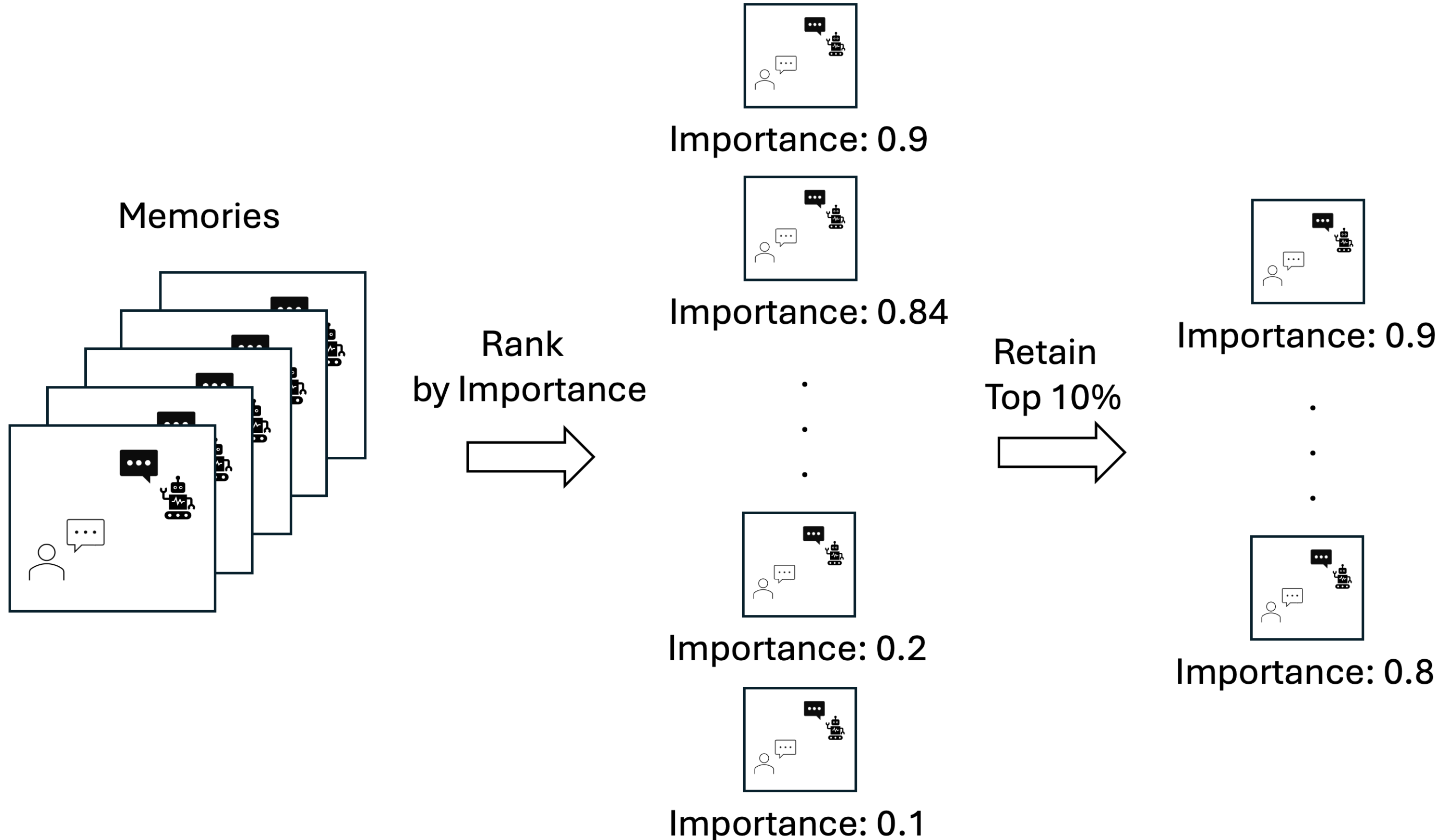}
\caption{Overview of the forgetting process. After a conversation ends, memories are ranked by computed importance, and only the top 10\% are retained.}
\label{fig: forgetting_process}
\end{figure}

\subsection{Forgetting Process}
\label{subsection:Forgetting Process}

As depicted in Figure~\ref{fig: forgetting_process}, the forgetting process is executed in two steps: ranking the memories according to importance, and retaining the top 10\% important memories. 
The forgetting process is executed only after the user finishes a conversation, ensuring that memory importance—calculated using retrieval frequency and other dynamic metrics—reflects the full interaction.

\paragraph{Retention Rate} As shown in Table~\ref{tab:RetentionStats}, MemoryBank and LUFY remembered a very small portion, mostly less than 10\% of the conversations. Here, each session refers to a 30-minute conversation between a human participant and the chatbot, as described in Section~\ref{sec: User Experiment} (see Section~\ref{subsubsec: Interaction Phase} for details).

Due to the discrete nature of memory chunks (i.e., both the total number of memories and the number of deleted items are whole numbers), the fraction of retained memories is not always exactly 10\%, though it remains consistently close. For instance, if there are 14 memories and 10\% corresponds to 1.4 items, we must round to the nearest whole number—e.g., retaining only 1 item. In this case, 1 out of 14 yields approximately 7.1\%, which is slightly lower than the intended 10\%.

\begin{table}[!ht]
    \centering
    \begin{tabular}{lcccc}
    \hline
    \textbf{System} & \textbf{S1} & \textbf{S2} & \textbf{S3} & \textbf{S4} \\
    \hline
    \multicolumn{5}{l}{\textit{Number of Retained Memories}} \\
    Naive RAG & 28.5 & 63.2 & 97.9 & 131.2 \\
    MemoryBank & 2.8 & 6.4 & 9.8 & 13.5 \\
    LUFY & 2.8 & 6.1 & 9.7 & 12.8 \\
    \hline
    \multicolumn{5}{l}{\textit{Retention Rate (\%)}} \\
    Naive RAG & 100 & 100 & 100 & 100 \\
    MemoryBank & 9.90 & 9.88 & 9.91 & 9.99 \\
    LUFY & 9.90 & 9.88 & 9.93 & 10.07 \\
    \hline
    \end{tabular}
    \vspace{10pt}
    \caption{Comparison of memory retention for Naive RAG, MemoryBank, and LUFY across four sessions. The table shows the cumulative number of memories retained (top) and the corresponding retention rate in percent (bottom).}
    \label{tab:RetentionStats}
\end{table}

\begin{table}[ht]
\setlength{\tabcolsep}{3mm}
    \centering
    \begin{tabular}{lcl}
    \hline
    \multicolumn{1}{c}{\textbf{System}}       & \textbf{Forgetting?} & \multicolumn{1}{c}{\textbf{Retrieval method}}          \\ \hline
    Naive RAG             & \texttimes                    & Cos. Sim.                  \\
    MemoryBank~\citep{zhong2024memorybank}         & \checkmark                    & Cos. Sim.\\
    LUFY           & \checkmark                    & Cos. Sim. + Importance
    \\ \hline
        \end{tabular}     
        \vspace{10pt}
    \caption{Comparison of systems in terms of forgetting and retrieval methods. \textbf{Naive RAG}: a conventional RAG chatbot model that exhibits no forgetting, \textbf{MemoryBank}: the model from previous work~\citep{zhong2024memorybank}, and \textbf{LUFY}: our proposed model.}
    \label{tab: system_comparison}
\end{table}

\section{User Experiment}
\label{sec: User Experiment}

We compared three different RAG chatbots: \textbf{Naive RAG}, \textbf{MemoryBank}, and \textbf{LUFY}. Naive RAG stores all memories, while MemoryBank and LUFY are equipped with a forgetting mechanism that retains only 10\% of memories. MemoryBank assesses the importance of a memory based solely on retrieval counts, whereas LUFY evaluates memory importance using six memory-related metrics, as depicted in Figure~\ref{fig: importance_assignment}. Additionally, although both Naive RAG and MemoryBank use cosine similarity only to assess the relevance of a memory to current conversations, LUFY also considers importance in this assessment. The differences between the three systems are summarized in Table~\ref{tab: system_comparison}. For an illustrative scenario highlighting how Naive RAG, MemoryBank, and LUFY differ in memory retention and retrieval behavior, see Appendix~\ref{appendix: Illustrative Example of Memory Retention and Retrieval Across Systems}.

Our study involved extended multi-day interactions that exceeded the length of prior dialog benchmarks~\citep{xu2022beyond}, as shown in Table~\ref{tab: Statistics compared to existing dialog datasets}.
\subsection{Procedure}

\subsubsection{Interaction Phase}
\label{subsubsec: Interaction Phase}
Seventeen participants each engaged in ten 30-minute conversations (hereafter referred to as  “sessions") over the course of 4 days. On Day~1, participants interacted with a Naive RAG chatbot---a standard RAG chatbot that uses cosine similarity to retrieve relevant documents based on query embeddings and has no forgetting process. The three chatbots---Naive RAG, MemoryBank, and LUFY---were all equipped with identical memories for this first session, although the memories underwent different forgetting mechanisms. From Day~2 to Day~4, participants interacted with all three chatbots for 30 minutes each, in a randomized order, to control for ordering effects. In total, there were four sessions for each chatbot. However, because the first session (Day~1) was identical for all three chatbots, participants engaged in a total of 10 sessions ($1 + 3 \times 3$).

The Day-1 chatbot was intentionally presented without a name. Participants were told only that they would have a 30-minute conversation with “a chatbot,” without any persona label or identifying information. This ensured that they understood they were interacting with a single, unnamed system on Day 1, rather than three separate bots. Beginning on Day 2, each of the three chatbots was assigned a distinct name, and participants were explicitly shown which named bot they were interacting with in each session. At the start of Day 2, participants were explicitly informed that all three named chatbots had access to the same Day-1 conversation history. They were also told that the systems used different underlying mechanisms, though the specific nature of these differences was not disclosed. This design made the three systems clearly distinguishable from Day 2 onward while maintaining a neutral, shared starting point on Day 1.

Each chatbot maintained a consistent identity across its four sessions, though none disclosed its underlying memory mechanism. Participants generally treated the three systems as separate interlocutors, aided by their distinct names and the divergent conversational trajectories that naturally emerged with each bot. On occasions when participants were unsure which bot they were speaking with—particularly when several days (and in a few cases more than a week) had elapsed between sessions—we provided a brief summary of their past conversations with that specific bot at the start of the session. These summaries served as gentle reminders of the bot’s prior interactions and identity, helping participants engage with each system as a distinct conversational partner with its own history.

\subsubsection{Post-Interaction Phase}
After each session, the participants were asked to create three question-and-answer (QA) pairs about the conversation they had just had. These questions contained information that the participants wanted the chatbot to remember about them. Additionally, the questions were designed to be clearly judged on a binary scale (1: correct, 0: incorrect) and to ensure that the answers would remain consistent in future interactions. If the participants' questions did not meet this criterion, they were asked to revise them. An example of a question that meets this criterion is 'What is my pet’s name?'

\subsection{LUFY-Dataset}
We developed a comprehensive dataset of human-chatbot conversations along with annotations for the important utterances, which we refer to as the \textbf{LUFY-Dataset}. 

Similar to the annotation procedure described in Section~\ref{sec: Quantifying Memory Importance}, at least three annotators reviewed and labeled the conversations from the user experiment using a binary scale: 1 for "important" and 0 for "unimportant." The Fleiss’ kappa score for inter-annotator agreement was 0.35, indicating fair agreement. In addition to the annotation of important memories, participants created three QA pairs for each conversation they had with the chatbot. Thus, the QA pairs directly correspond to their respective conversations. Examples are illustrated in Table~\ref{tab: Example conversation of the LUFY-Dataset} and \ref{tab: Example QA pairs of the LUFY-Dataset}.

For the de-identification process, multiple reviewers examined the conversations to ensure that all personally identifiable information (PII) was removed or modified. 

\paragraph{Removal of Direct Identifiers} This involves eliminating information that can directly identify an individual, such as names, phone numbers, and addresses.

\paragraph{Generalization} Specific data points are replaced with broader categories to balance privacy and data utility. For instance, exact ages might be replaced with age ranges (e.g., “29” becomes “20–30”), or full dates (e.g., “April 15, 1995”) might be replaced with just the year (“1995”). This process helps prevent re-identification, especially when combined with other quasi-identifiers. The extent of generalization depends on the dataset's intended use. For datasets used in model training or longitudinal analysis, overly aggressive generalization might reduce usefulness. In contrast, applications with stricter privacy requirements may necessitate more coarse-grained representations. In our case, since the goal was to release a publicly available benchmark while preserving meaningful conversational structure and user traits, we opted for moderate generalization, preserving categories like profession or age range where appropriate.

\begin{table}[ht]
\centering
\begin{tabular}{|c|c|l|}
\hline
\textbf{Important?} & \textbf{Speaker} & \textbf{Utterance} \\ \hline
0 & User & Hey, how are you today? \\ \hline
0 & Bot & I’m doing well, thanks! What’s up? \\ \hline
1 & User & I just found out I got into Harvard! \\ \hline
0 & Bot & Wow, congratulations! That’s amazing! \\ \hline
0 & User & Thanks, I’m still processing it all. \\ \hline
\multicolumn{3}{|c|}{\textellipsis\ \textit{conversation continues}\ \textellipsis} \\ \hline
\end{tabular}
\vspace{10pt}
\caption{Example conversation of the LUFY-Dataset}
\label{tab: Example conversation of the LUFY-Dataset}
\end{table}

\begin{table}[ht]
\centering
\begin{tabular}{|c|l|}
\hline
Q1 & Which university did \{user\} get accepted to?\\ \hline
A1 & Harvard. \\ \hline
Q2 & What is \{user\}'s favorite dog breed? \\ \hline
A2 & White Schnauzer. \\ \hline
Q3 & What is the name of \{user\}'s dog? \\ \hline
A3 & Luke. \\ \hline
\end{tabular}
\vspace{10pt}
\caption{Example QA pairs of the LUFY-Dataset}
\label{tab: Example QA pairs of the LUFY-Dataset}
\end{table}

Some studies, such as DeID-GPT~\citep{liu2023deid}, explore the use of LLMs for de-identification. Other studies, such as the Ego4D dataset~\citep{grauman2022ego4d}, which consists of 3,000 hours of video, use commercial software like Brighter.ai and SecureRedact. However, we manually executed the process to ensure proper de-identification.

We aim for the \textbf{LUFY-Dataset} to serve as a benchmark for long-term conversations, given its unique length, as shown in Table~\ref{tab: Statistics compared to existing dialog datasets}. The full dataset, including both the annotated human-chatbot conversations and the participant-created QA pairs, is publicly released to support future research in long-term conversational AI.




\subsection{Main Results}

Firstly, we evaluated the user experience using both subjective and automatic methods. We used three evaluation methods: (1) \textit{subjective human ratings} (a subjective method) and (2) \textit{LLM-based scoring} (an automatic method), both based on the \textit{entire conversation}, and (3) \textit{sentiment analysis} (an automatic method), which was performed at the \textit{individual user utterance level}.

\paragraph{Subjective Evaluations} For subjective evaluation, we asked third-party annotators to rate each conversation on three criteria: personalization, flow of conversation, and overall experience. To ensure that personalization judgments reflected the evolving relationship between each participant and chatbot, annotators evaluated the conversations in chronological order per participant, giving them access to earlier sessions when rating later ones. This ordering allowed raters to apply criteria consistently across multi-day interactions rather than evaluating each session in isolation. We provide the specific instructions for the three criteria, together with the Intraclass Correlation Coefficient (ICC)~\citep{shrout1979intraclass}, in Appendix~\ref{appendix: instructions for annotators for subjective eval}.

We opted for third-party annotators instead of collecting user self-ratings for several reasons. While user opinions are ultimately the most valuable signal, we were concerned that prompting users to rate each chatbot after every session might introduce bias—participants could infer which system was expected to perform better, potentially influencing their subsequent behavior. Additionally, participants engaged in multiple long sessions (approximately five hours across four days); frequent questionnaires could contribute to fatigue and degrade the quality of feedback. To avoid these issues and ensure consistency across systems, we relied on independent annotators.

The results, as presented in Table~\ref{tab: Personalization, Flow of Conv., Overall}, demonstrate LUFY's superior performance over both Naive RAG and MemoryBank across most sessions and evaluation criteria, with the most significant difference observed in Session 4, where ratings for Naive RAG and MemoryBank dropped significantly, while LUFY's rating remained relatively stable. This trend underscores LUFY's enhanced ability to maintain engaging and personalized conversations, particularly as the interaction length increases.

\begin{table*}[!ht]
\centering
\setlength{\tabcolsep}{1.7pt}
\renewcommand{\arraystretch}{1.2}
\begin{tabular}{lccccccccccccccccc}
\hline
& \multicolumn{5}{c}{Personalization} & & \multicolumn{5}{c}{Flow of Conv.} & & \multicolumn{5}{c}{Overall} \\
System & S1 & S2 & S3 & S4 & Avg. & & S1 & S2 & S3 & S4 & Avg. & & S1 & S2 & S3 & S4 & Avg.\\ \hline
Naive RAG & (3.94) & 3.85 & 3.89 & 3.76 & 3.83 & & (3.86) & 3.46 & 3.57 & 3.20 & 3.41 & & (3.82) & 3.65 & \textbf{3.74} & 3.50 & 3.63 \\
MemoryBank & (3.87) & 3.95 & 3.87 & 3.56 & 3.79 & & (3.69) & 3.58 & \textbf{3.64} & 3.31 & 3.51 & & (3.85) & 3.75 & 3.64 & 3.36 & 3.58 \\
\rowcolor{gray!20}
LUFY & (3.94) & \textbf{4.13}$^\dagger$  & \textbf{3.98} & \textbf{4.04}$^\dagger$  & \textbf{4.05}$^\dagger$  & & (3.78) & \textbf{3.63} & 3.57 & \textbf{3.72}$^\dagger$  & \textbf{3.64}$^\dagger$  & & (3.80) & \textbf{3.81} & \textbf{3.74} & \textbf{3.80}$^\dagger$  & \textbf{3.78}$^\dagger$  \\ \hline
\end{tabular}
\caption{Subjective Evaluation of three criteria: Personalization, Flow of Conversation and Overall, with ratings on a scale from 1 (lowest) to 5 (highest). $\dagger$ indicates statistically significant improvement over both other methods.($p < 0.05$
, paired t-test). }
\label{tab: Personalization, Flow of Conv., Overall}
\end{table*}

\paragraph{Rating by LLM} 

As part of the automatic method to score the user experience, we assessed the user's overall satisfaction using GPT-4o. We used GPT-4o as an evaluator because strong LLMs achieve performance comparable to human evaluators~\citep{zheng2024judging}. The prompt given to the user experience is given in Figure~\ref{fig:prompt for estimation of User Experience}. As shown in Table~\ref{tab: Human Evaluation by GPT-4o}, the results provide valuable insights into the performance of the three RAG chatbots. LUFY consistently emerges as the top performer, achieving the highest average rating and demonstrating a significant improvement in the later stages of the conversation (Session 4). This aligns with the subjective evaluations and is further supported by the sentiment analysis in the following section.

\begin{table}[t]
    \centering
    \renewcommand{\arraystretch}{1.2}
    \setlength{\tabcolsep}{4pt}
    \begin{tabular}{lccccc}
    \hline
    System & (S1) & S2 & S3 & S4 & Avg. \\
    \hline
    Naive RAG & (3.71) & \textbf{3.53} & 3.38 & 3.18 & 3.36 \\
    MemoryBank & (3.71) & \textbf{3.53} & \textbf{3.44} & 3.21 & 3.39 \\
    \rowcolor{gray!20}
    LUFY & (3.71) & \textbf{3.53} & 3.32 & \textbf{3.50}$^\dagger$ & \textbf{3.45} \\
    \hline
    \end{tabular}
    \vspace{10pt}
    \caption{
        Average ratings (1–5 scale) by GPT-4o, averaged over 30 ratings. 
        $\dagger$ indicates statistically significant improvement over both other methods.($p < 0.05$
, paired t-test). 
        Ratings are shown with mean values.
    }
    \label{tab: Human Evaluation by GPT-4o}
\end{table}

\begin{figure}[ht]
\centering
\begingroup
\footnotesize  
\begin{verbatim}
Based on the following dialogue, evaluate and score 
how engaging the user({user_name})'s conversation is.

Consider the following factors: 
1. the relevance of responses
2. ability to sustain a conversation
3. interest generated through their responses

Here is the conversation script:{conversation_script}

The score should be on a scale from 1 to 5,
where 1 is the lowest and 5 is the highest. 
Be sure to provide both a score and a brief reason for your rating.

\end{verbatim}
\endgroup
\caption{Prompt for estimation of User Experience.}
\label{fig:prompt for estimation of User Experience}
\end{figure}

\paragraph{Sentiment Analysis} 

To complement both subjective and LLM-based evaluations, we also performed sentiment analysis at the utterance level. We conducted sentiment analysis using fine-tuned versions of RoBERTa models. We used DistilRoBERTa~\citep{sanh2019distilbert}, fine-tuned on 4,840 polar sentiment sentences from English financial news, and TimeLMs~\citep{loureiro-etal-2022-timelms}, a RoBERTa model trained on 124M tweets from January 2018 to December 2021 and fine-tuned with the TweetEval benchmark~\citep{mohammad2018semeval}. The hyperparameters for the fine-tuning of DistilRoBERTa are listed in Table~\ref{tab: Hyperparameters for distilRoBERTa}.

\begin{table}[ht]
\centering
\begin{tabular}{@{}ll@{}}
\toprule
\textbf{Parameter} & \textbf{Value} \\
\midrule
Learning Rate & 2e-05 \\
Train Batch Size & 8 \\
Eval Batch Size & 8 \\
Seed & 42 \\
Optimizer & Adam \\
\quad $\beta_1, \beta_2 $& 0.9, 0.999 \\
\quad Epsilon & 1e-08 \\
LR Scheduler Type & Linear \\
Number of Epochs & 5 \\
\bottomrule
    \end{tabular}     
    \vspace{10pt}
\caption{Hyperparameters Used for Fine-tuning of DistilRoBERTa}
\label{tab: Hyperparameters for distilRoBERTa}
\end{table}

\begin{table}[!ht]
    \centering
    \setlength{\tabcolsep}{4pt}
    \renewcommand{\arraystretch}{1.2}
    \begin{tabular}{@{}lccccc@{}}
    \toprule
        System & S1 & S2 & S3 & S4 & Avg. \\
    \midrule
        Naive RAG & (0.58) & \textbf{0.51} & 0.38 & 0.32 & 0.40 \\ 
        MemoryBank& (0.58) & 0.47 & 0.34 & 0.30 & 0.38 \\
        \rowcolor{gray!20}
        LUFY & (0.58) & 0.43 & \textbf{0.45} & \textbf{0.62}$^\dagger$ & \textbf{0.50}$^\dagger$ \\
    \bottomrule
        \end{tabular}     
        \vspace{10pt}
    \caption{Sentiment Analysis Results: Average rating of user utterances (+1 = positive, 0 = neutral, -1 = negative). $\dagger$ indicates statistically significant improvement over both other methods.}
    \label{tab: Sentiment Analysis}
\end{table}

\begin{table*}[!ht]
    \centering
    \renewcommand{\arraystretch}{1.2}
    \setlength{\tabcolsep}{2.2pt}
    \begin{tabular}{lcccccccccccccccccc}
    \hline
    & \multicolumn{5}{c}{Precision} & & \multicolumn{5}{c}{Recall} & & \multicolumn{5}{c}{F1 Score} \\
    System & S1 & S2 & S3 & S4 & Avg. & & S1 & S2 & S3 & S4 & Avg. & & S1 & S2 & S3 & S4 & Avg. \\ \hline
    Naive RAG & 75.6 & 69.6 & 74.2 & 69.9 & 72.3 & & \textbf{60.8} & \textbf{51.0} & \textbf{51.6} & \textbf{46.6 }& \textbf{52.5} & & \textbf{67.4} & 58.9 & \textbf{60.9} & \textbf{55.9} & \textbf{60.8} \\
    MemoryBank & 63.6 & 50.7 & 57.4 & 64.6 & 59.1 & & 41.2 & 29.4 & 30.7 & 31.9 & 33.3 & & 50.0 & 37.2 & 40.0 & 42.7 & 42.5 \\
    \rowcolor{gray!20}
    LUFY & \textbf{80.0} & \textbf{86.9} & \textbf{86.8} & \textbf{86.6} & \textbf{85.1} & & 54.9 & 51.0 & 38.6 & 35.3 & 44.9 & & 65.1 & \textbf{64.3} & 53.4 & 50.2 & 58.3 \\ \hline
    \end{tabular}
    \caption{Comparison of Precision, Recall, and F1 Score for different systems and sessions. Each value reflects performance on the full set of questions encountered up to that session. For example, Recall in S3 measures the model’s ability to correctly answer questions from sessions 1 through 3 at the end of session 3.}
    \label{tab: PrecisionRecallF1}
\end{table*}

As shown in Table~\ref{tab: Sentiment Analysis}, the sentiment analysis results clearly indicate that LUFY outperformed both Naive RAG and MemoryBank in fostering positive user experiences, especially during longer interactions. LUFY consistently achieved higher average sentiment scores (0.50), demonstrating its ability to maintain a positive conversational tone, particularly evident in Session 4 where its score (0.62) significantly surpassed those of Naive RAG and MemoryBank. This trend aligns with the subjective evaluations, highlighting its effectiveness in extended conversational settings.

While sentiment analysis provides valuable insights into the emotional tone of user utterances, we acknowledge that it does not capture the full nuance of user experience. In particular, conversations involving argumentation or critical reflection may contain negative sentiment while still being engaging, constructive, and intellectually stimulating. Therefore, we do not treat sentiment scores as a standalone measure of user satisfaction or engagement. Instead, we use sentiment analysis as a complementary signal to the subjective human evaluations and LLM-based scoring, which are better suited to capture the broader context and qualitative aspects of the interaction.

\paragraph{Summary of Results}

In summary, the subjective evaluations, LLM-based ratings, and sentiment analysis provide a consistent picture: LUFY excels in delivering sustained, personalized, and engaging user experiences, particularly in longer interactions. Across all three evaluation methods, LUFY achieved the highest scores, most notably in Session 4, where both Naive RAG and MemoryBank exhibited significant performance drops. In contrast, LUFY maintained stable and positive engagement. While Naive RAG and MemoryBank performed comparably to each other, their reliance on cosine similarity for memory retrieval and the limitations of MemoryBank's forgetting mechanism appear to limit their ability to support coherent long-form conversations. These findings strongly suggest that LUFY's use of an importance-based forgetting mechanism, combined with importance-aware memory retrieval, is effective for improving chatbot quality in extended interactions.

\subsection{In-depth Analysis}
\label{sec: In-depth Analysis}


\paragraph{Evaluations with QA pairs}
We assessed the models' ability to remember using the QA pairs collected in the User Experiment. After each session—and, for MemoryBank and LUFY, also after their respective forgetting processes (Naive RAG does not include a forgetting process)— we asked each model the full set of questions accumulated up to that point and calculated Precision, Recall, and F1 Score based on its responses.

To ensure consistency and objectivity, responses were scored by GPT-4o, an LLM. The exact prompt used for evaluation is detailed in Appendix~\ref{appendix: Prompt for judging whether the response is correct or not}. Summary results are presented in Table~\ref{tab: PrecisionRecallF1}, while the full set of results is available in Appendix~\ref{appendix: Complete Results}.

Naive RAG achieved the highest recall overall, because of its lack of a forgetting mechanism. However, this came at the cost of lower precision compared to LUFY, suggesting that it retrieved more irrelevant or outdated information. LUFY outperformed both Naive RAG and MemoryBank in Precision, and surpassed MemoryBank across all metrics in all sessions, indicating a more effective method of measuring importance than MemoryBank.

\begin{table}[t]
    \centering
    \renewcommand{\arraystretch}{1.2}
    \setlength{\tabcolsep}{3pt}
    \begin{tabular}{lccccccc}
    \hline
    & \multicolumn{5}{c}{Important Memories} \\
    System & S1 & S2 & S3 & S4 & Avg.\\ \hline
MemoryBank & \textbf{19.4} & 8.8 & 14.3 & \textbf{15.3} & 14.4 \\
\rowcolor{gray!20}
LUFY & \textbf{19.4} & \textbf{16.3} & \textbf{24.4} & 10.1 & \textbf{17.6} \\
(Random) & 13.1 & 10.0 & 10.6 & 11.1 & 11.2 \\
(Annotators) & 18.3 & 26.1 & 23.5 & 24.2 & 23.0 \\
 \hline
        \end{tabular}     \vspace{10pt}
    \caption{System–Human Agreement on Memory Importance.}
    \label{tab: Agreement probability}
\end{table}

\begin{table*}[!ht]
    \centering
    \renewcommand{\arraystretch}{1.2}
    \setlength{\tabcolsep}{3pt}
    \begin{tabular}{lcccccc@{\hspace{8pt}}cccccc@{\hspace{8pt}}ccccc}
    \hline
    & \multicolumn{5}{c}{Precision} & & \multicolumn{5}{c}{Recall} & & \multicolumn{5}{c}{F1 Score} \\
    System & S1 & S2 & S3 & S4 & Avg. & & S1 & S2 & S3 & S4 & Avg. & & S1 & S2 & S3 & S4 & Avg. \\ \hline
    LUFY & 80.0 & 86.9 & 86.8 & 86.6 & 85.1 & & 54.9 & 51.0 & 38.6 & 35.3 & 45.0 & & 65.1 & 64.3 & 53.4 & 50.2 & 58.3 \\
    $-A$ & \underline{64.3} & 84.3 & 81.4 & 82.2 & \underline{78.1} & & \underline{37.3} & 52.0 & 33.4 & 32.9 & 38.9 & & \underline{47.2} & 64.3 & 47.4 & 47.0 & \underline{51.5} \\
    $-P$ & 79.0 & 84.3 & \underline{77.3} & 80.2 & 80.2 & & 45.1 & 45.1 & \underline{32.1} & 29.4 & \underline{38.0} & & 57.4 & 58.8 & \underline{45.4} & 43.0 & 51.7 \\
    $-L$ & 74.8 & 83.8 & 83.2 & \underline{77.1} & 79.7 & & 49.0 & \underline{44.1} & 32.8 & \underline{28.9} & 38.7 & & 59.2 & \underline{57.8} & 47.1 & \underline{42.0} & \underline{51.5} \\
    $-R1$ & 68.6 & \underline{82.0} & 82.3 & 87.0 & 80.0 & & 47.1 & 47.1 & 38.0 & 38.7 & 42.7 & & 55.9 & 59.8 & 52.0 & 53.6 & 55.3 \\
    $-R2$ & 80.0 & 86.9 & 86.8 & 86.6 & 85.1 & & 54.9 & 51.0 & 38.6 & 35.3 & 45.0 & & 65.1 & 64.3 & 53.4 & 50.2 & 58.3 \\
    \hline
    \end{tabular}
    \caption{Ablation Study for Precision, Recall, and F1 Score. $-A$ indicates that $w_A$ is set to 0. The largest drop in performance is underlined.}
    \label{tab: Ablation Study}
\end{table*}

\paragraph{Memory matching rate}
To evaluate how closely system-selected memories align with human judgments, we measure the average pairwise agreement between each system and individual human annotators. For each session, we compare the binary importance labels produced by the model with those assigned by each annotator, and compute the proportion of system-selected memories that were also marked as important by that annotator. 

As shown in Table~\ref{tab: Agreement probability}, across sessions, LUFY outperforms or matches MemoryBank in three out of four sessions and achieves a higher overall average agreement, demonstrating more consistent alignment with human judgments. Notably, LUFY even exceeds the human–human pairwise agreement in two sessions, although its overall mean still falls slightly below the human average—reflecting the substantial subjectivity inherent in memory-importance annotation, where human pairwise agreement is only around 23\%.

We also computed the precision over utterances unanimously marked as important by all annotators. However, such unanimously important cases were extremely rare (typically only 2–6 per session), making the resulting precision estimates highly unstable and therefore unsuitable as a primary evaluation metric. For this reason, we focus on the more reliable pairwise agreement analysis.

Agreement on unimportant memories (i.e., cases where both the system and humans marked an utterance as unimportant) is much higher due to class imbalance—approximately 90\% of utterances receive a “not important” label—but provides little discriminatory value between systems and is therefore omitted.

\paragraph{Ablation Study}
We conducted an ablation study for the agreement probability with other annotators and precision, recall and F1 Score. As shown in Tabls~
\ref{tab: Ablation Study} and~\ref{tab: Ablation Study for Agreement probability}, we found A (Arousal), L (LLM estimated importance), R1 (the number of times the memory is retrieved) to be of particular importance.

\begin{table}[t]
    \centering
    \renewcommand{\arraystretch}{1.2}
    \setlength{\tabcolsep}{3pt}
    \begin{tabular}{lcccccc}
    \hline
    & \multicolumn{5}{c}{Important Memories} \\
    System & S1 & S2 & S3 & S4 & Avg.\\ \hline
    LUFY & 20.9 & 12.4 & 11.8 & 14.4 & 14.9 \\
    $-A$ & 21.5 & 11.8 & \underline{9.1} & 14.4 & \underline{14.2} \\
    $-P$ & 23.5 & 11.8 & 10.4 & \underline{13.7} & 14.9 \\
    $-L$ & \underline{20.2} & 11.8 & 9.8 & 15.0 & \underline{14.2} \\
    $-R1$ & 21.5 & \underline{10.4} & 11.1 & \underline{13.7} & \underline{14.2} \\
    $-R2$ & 20.9 & 12.4 & 11.8 & 14.4 & 14.9 \\
    \hline
        \end{tabular}     \vspace{10pt}
    \caption{Ablation Study for the agreement probability with other annotators on whether memories are important. The largest drop in performance is underlined.}
    \label{tab: Ablation Study for Agreement probability}
\end{table}

\paragraph{Episodic Memory in Conversations}
As stated in Section~\ref{subsubsec: Retrieval Method}, we stored three types of information: profile information, summaries of past sessions, and utterance pairs. We include summaries and utterance pairs in memory storage to retain small details about the user. These details, such as episodic memories~\citep{tulving2002episodic}, are unlikely to appear in the profile information.

To assess the necessity of retaining such details, we measured the prevalence of episode-related utterances—user statements referring to unique, personally experienced events. We used an LLM to identify such utterances. Specifically, the model was given the prompt shown in Figure~\ref{fig: Prompt template for identifying whether a user utterance pertains to an episodic memory.}:

This classification process was applied across a sample of 170 conversations. Our analysis revealed that 10.6\% of the user's utterances were episode-related. As shown in Figure~\ref{fig: pie_chart}, only 36 of the 170 analyzed 30-minute conversations (21.1\%) did not contain any episode-related utterances. In some conversations, however, up to 40\% of the utterances were classified as episode-related.

\begin{figure}[ht]
\centering
\begingroup
\footnotesize  
\begin{verbatim}
Identify any parts of this conversation where the speaker refers 
to specific past experiences or events they personally remember. 

These should not include general knowledge, facts, or information 
from their profile, but rather unique, one-time occurrences or 
episodic memories. 

If no such references are found, output ‘0’. 
Here is the conversation: \{conversation\}  

\end{verbatim}
\endgroup
\caption{Prompt template for identifying whether a user utterance pertains to an episodic memory.}
\label{fig: Prompt template for identifying whether a user utterance pertains to an episodic memory.}
\end{figure}

\begin{figure}[htbp]
  \centering
  \scalebox{1}{
    \includegraphics[width=\linewidth]{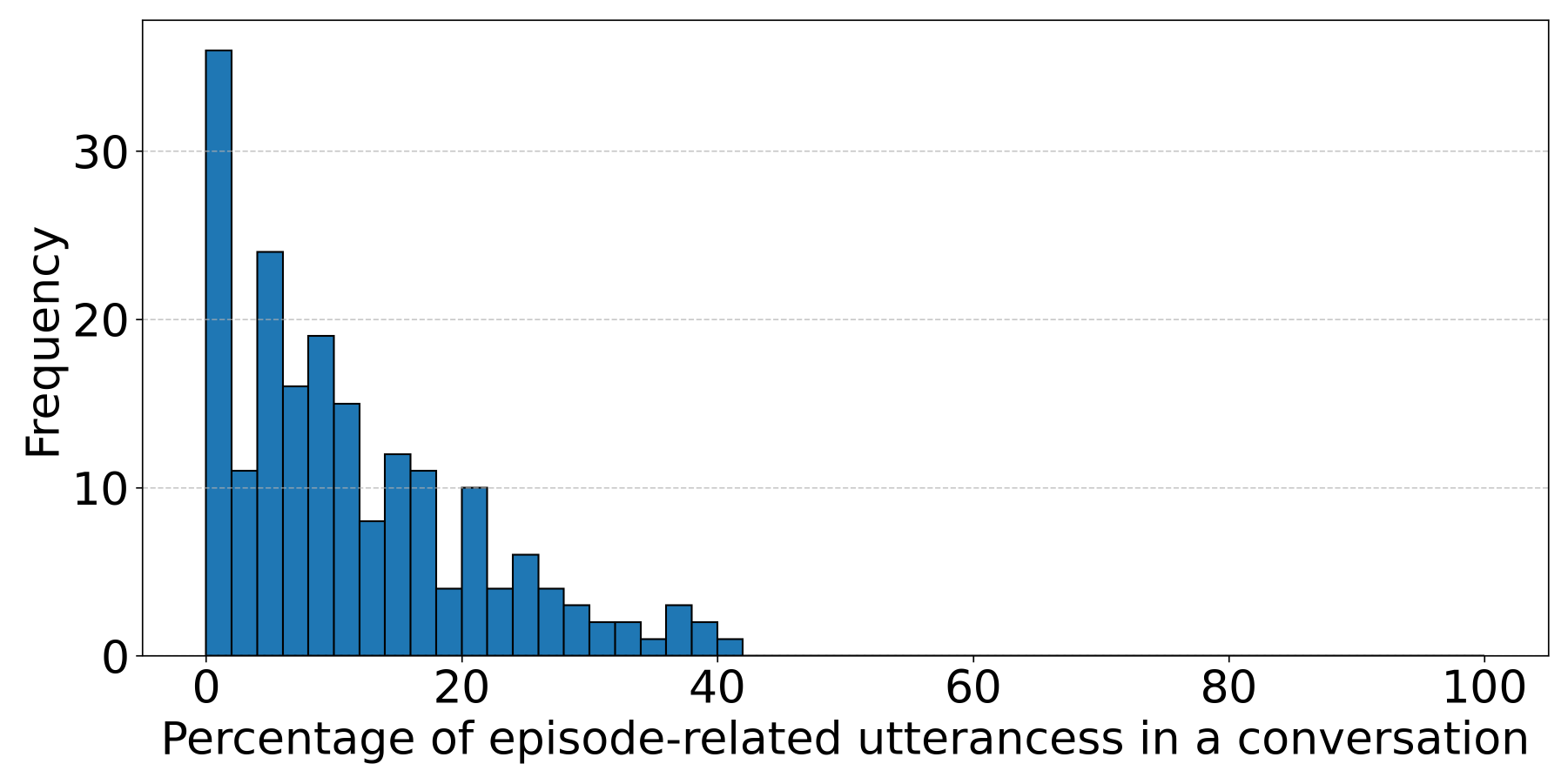}
  }
  \caption{Episode-related utterances frequency distribution in conversations (N = 170)}
  \label{fig: pie_chart}
\end{figure}

\section{Related Work}
\label{sec: Related Work}

The evolution of conversational AI from stateless chatbots to modern LLMs has been defined by an increasing capacity for memory. While the Transformer architecture’s context window provides a form of short-term memory, enabling conversational coherence, it is insufficient for the long-term, personalized interactions that characterize human relationships. To achieve this, agents require persistent, structured memory architectures capable of recalling specific interactions (episodic memory) and accessing a stable base of world knowledge (semantic memory). This review connects foundational theories from cognitive psychology with contemporary computational models, arguing that the future of conversational intelligence lies in the thoughtful integration of these cognitively-inspired memory systems.

Our work is situated at the intersection of computational dialogue modeling and cognitive memory theory. Foundational models in cognitive psychology distinguish between short-term and long-term memory~\citep{atkinson1968human}, and further between semantic memory (general knowledge) and episodic memory (event-specific personal experiences)~\citep{tulving1972episodic}. Episodic memory, in particular, is crucial in dialogue, where references to past interactions and experiences often shape conversational flow and user engagement. Our analysis of the LUFY corpus found that approximately 79\% of conversations included episode-related utterances, underscoring the importance of modeling episodic recall in conversational agents.

Traditional memory models in dialogue systems often rely on salience mechanisms such as recency and frequency (as in MemoryBank). While these approaches are computationally efficient, they fail to capture the richness of human remembering and forgetting, which involve not just decay and interference but also selective omission and social reasoning. In dialogue, speakers may choose to withhold or imply information based on context, face-saving~\citep{brockner1981face}), or relationship management—processes not easily modeled with surface-level heuristics alone.

To address these challenges, recent work has explored structured and cognitively motivated memory systems. Garcia Contreras et al.~\citep{garcia2024forgetful} introduced a \textit{multi-store memory framework} for an assistive robot, Indy, inspired by cognitive psychology. Their system is built around a three-tiered architecture—sensor memory, short-term memory (STM), and long-term memory (LTM)—and incorporates forgetting heuristics based on parametrized Weibull decay curves~\citep{murthy2004weibull}. These mechanisms enable the robot to retain only contextually relevant or repeatedly reinforced information, simulating aspects of human episodic memory without unbounded memory growth.

Our current system retrieves past conversational content using a RAG framework. While this supports some degree of contextual continuity, it does not yet implement explicit mechanisms for episodic segmentation or narrative abstraction as described in Garcia Contreras et al.'s work. Their multi-store memory framework, built around sensor, short-term, and long-term memory, introduces cognitively inspired forgetting heuristics and structured narrative memory. Recent work by~\citep{ong2025towards} also highlights the importance of structuring prior conversational data, proposing a memory timeline framework (THEANINE) that links past memories through temporal and causal relations to support response generation. While distinct from our goals, such efforts underscore the broader value of organizing conversational history beyond flat retrieval. In addition, we see promising avenues for future enhancement through structured relational representations. Incorporating knowledge graphs or narrative schemas~\citep{wilcock2022conversational, walker2022dialogue} would enable the system to construct and retrieve memories not just as isolated utterances, but as semantically coherent entities and events—mirroring how humans organize and recall lived experience. 

As discussed above, while LUFY already demonstrates the effectiveness of cognitively informed forgetting and memory prioritization, its design also opens up natural extensions toward structured and relational memory—marking a promising direction for future development.


\section{Conclusion}
\label{sec: Conclusion}

\subsection{Summary}
This study introduced LUFY, a novel Retrieval-Augmented Generation (RAG) chatbot designed to address the challenge of memory management in long-term conversations. Drawing on psychological insights, LUFY employs a unique approach to evaluate and prioritize memories based on six key metrics: emotional arousal, surprise, LLM-estimated importance, and retrieval-induced forgetting. Unlike traditional RAG models that either store all memories or rely on simplistic forgetting mechanisms, LUFY uses learned weights to balance these metrics, ensuring that emotionally significant and contextually relevant memories are retained while less important ones are gradually forgotten. Through extensive user experiments, involving conversations 4.5 times longer than those in previous studies, LUFY demonstrated superior performance in maintaining engaging, personalized, and positive dialogues. The results, validated through subjective evaluations, sentiment analysis, and LLM assessments, showed that LUFY significantly outperformed both a naive RAG system and the MemoryBank model, particularly in later stages of extended interactions. The development of the LUFY-Dataset, a comprehensive collection of human-chatbot conversations annotated for importance, further contributes to the field by providing a valuable resource for future research in long-term conversational AI. Overall, this study demonstrates the potential of forgetting unimportant memories.

\subsection{Future Work}
This study has demonstrated the potential advantages of the proposed method in enhancing chatbot interactions through improved objective and subjective evaluations. However, it is important to acknowledge its limitations.

Firstly, our investigation focused solely on the impact of memory-related psychological metrics in conversations between strangers. Future research should aim to diversify the dataset by including interactions among friends, family members, and other relationships to comprehensively understand these metrics' influence across different conversational contexts.

Second, this study only looked at text-based conversations. In text, indicators like exclamation marks play a role in detecting arousal. For example, when a user said, "I'm going to Hong Kong with my friends next month," our system recognized a low arousal level, missing the excitement. Future work should explore multimodal settings to improve the accuracy of emotional detection.

Third, our current framework does not yet capture implicitly referenced content, social dynamics (e.g., face-saving~\citep{brockner1981face}), or dialogue progression structure~\citep{levinson1981some} (e.g., adjacency pairs, repairs). Future work could investigate how such relational and pragmatic elements impact memory importance, potentially by incorporating discourse parsing, speaker state tracking, or theory-of-mind modeling. In addition, we plan to explore complementary metrics such as conversational risk-taking (e.g., controversy), and topical diversity to capture broader aspects of engagement beyond flow and personalization.

Fourth, in practice, long-term interactions may introduce contradictions—for example, when a user updates or retracts earlier statements. LUFY does not currently attempt contradiction detection at the symbolic level, but it mitigates conflict through dynamic forgetting and salience-based retrieval: newer, more emotionally charged, or frequently retrieved memories are prioritized, while outdated or less important ones are forgotten. This approach is consistent with prior work in long-term memory for social agents~\citep{bae2022keep}, which similarly favors overwriting older memories implicitly during retrieval. Future work could integrate more explicit contradiction resolution using techniques from natural language inference or entailment.

Fifth, while we standardized the prompts used for tasks such as memory importance estimation and user experience evaluation, we recognize that prompt design can meaningfully influence LLM behavior. For instance, in the rating prompt shown in Figure~\ref{fig: Prompt for LLM-estimated importance}, the numerical scale appears after the instructions. We did not investigate whether reordering or rephrasing such elements would alter outcomes. Future work could examine prompt sensitivity more systematically—for example, testing whether presenting the rating scale before vs. after the task description impacts consistency or alignment with human judgments. This direction may be especially important for ensuring the reliability and interpretability of LLM-based evaluations.

\paragraph{Ethical Considerations.} As our system involves the long-term retention of potentially sensitive user utterances---some of which are emotionally charged or self-disclosive---it raises important ethical questions for future deployment. In particular, ensuring user consent for what is stored, providing mechanisms for inspecting or deleting remembered content, and minimizing the risk of misrepresentation or inappropriate persistence of private information are essential. While our experiments were conducted with informed, consenting participants and involved de-identified data, practical interactive applications must prioritize transparency and user agency in memory handling. We envision future iterations of LUFY supporting user-controllable memory (e.g., editable memory logs or consent-based retention) to better align with ethical standards for human--AI interaction.



\bibliography{references}

\appendix

\section{Top-\textit{k} Retrieval}
\label{appendix: Top-k Retrieval}

\begin{table}[ht]
\centering
\begin{tabular}{lr}
\toprule
Top-\textit{k} Memory &  Ratio (\%) \\
\midrule
Top 1 &          52.5 \\
Top 2 &          61.8 \\
Top 3 &          71.6 \\
Top 4 &          79.9 \\
Top 5 &          84.8 \\
Top 6 &          85.6 \\
Top 7 &          88.2 \\
Top 8 &          88.2 \\
Top 9 &          90.6 \\
Top 10 &         91.2 \\
(Not in Top 10) & 8.8 \\
\bottomrule
    \end{tabular}     \vspace{10pt}
\caption{Ratio of the correct memory being in various top-\textit{k} memories. The correct memory is included in the top-5 memories 84.8\% of the time.}
\label{fig: Top-k}
\end{table}

Using the same 495 QA pairs from the LUFY-Dataset, we retrieved the top 10 memories for each question and analyzed the position of the correct memory among them. The results are shown in Table~\ref{fig: Top-k}. Notably, we found that 85\% of the correct memories appear within the top 5, with only marginal improvement when increasing \textit{k}. This suggests that selecting the top 5 memories is effective for conversational RAG chatbots, aligning with previous studies on RAG systems in non-conversational settings~\citep{li2024retrieval}. Aligning our method with the concept of RIF~\citep{hirst2012remembering}, we retrieved the top 2 memories despite the higher effectiveness observed with the top 5.

\section{Empirical Estimation of Contemporary Event References in the LUFY Corpus}
\label{appendix:contemporary}

To evaluate the reliability of parameter values grounded in literature estimates—particularly the claim that approximately 70\% of conversation time concerns contemporary events—we conducted an empirical analysis on a random sample of the LUFY dialogue corpus. Specifically, we analyzed 2,092 randomly selected utterances. Of these, 479 (22.90\%) concerned contemporary events, while 1,613 (77.10\%) did not, suggesting a considerably lower proportion of contemporary-event-focused dialogue than the literature estimate.

\subsection{Method}
We used an LLM to classify utterances according to whether they referenced contemporary or recent events. Specifically, the model was prompted with user-bot dialogue pairs and asked to identify whether the user's utterance referred to an event that is currently happening, just occurred, or is scheduled to occur soon. Contemporary utterances included examples such as “I’m going to the beach tomorrow” or “We’re watching the game tonight.” Non-contemporary utterances involved background information, personality traits, opinions, or general facts, such as “I love jazz” or “My sister is a doctor.”

Each utterance was independently evaluated using the prompt in Figure~\ref{fig: Prompt template for identifying whether a user utterance pertains to a contemporary or recent event.}.

\begin{figure}[ht]
\centering
\begin{verbatim}
You will be given a pair of utterances from a conversation: 
one from a user and one from a chatbot.

Determine whether the user’s utterance refers to a contemporary 
or recent event—that is, something happening now, 
that just happened, or that is planned to happen soon 
(e.g., "I'm going to the beach tomorrow", 
"I just got a new job", "We're watching the game tonight").

If the utterance is about general facts, opinions, 
personality traits, or background information 
(e.g., "I love jazz", "My sister is a doctor", "I’m a morning person"), 
it is not considered contemporary.

Here is the bot and user’s utterance:  
Bot: \{bot\_utterance\}  
User: \{user\_utterance\}

Output 1 if the user’s utterance is about contemporary things, 
0 if not. Do not output anything else—just the number.

\end{verbatim}
\caption{Prompt template for identifying whether a user utterance pertains to a contemporary or recent event.}
\label{fig: Prompt template for identifying whether a user utterance pertains to a contemporary or recent event.}
\end{figure}

\subsection{Results}
The analysis yielded the following statistics over a total of 2,092 user utterances from the LUFY corpus:

\begin{itemize}
    \item Total number of utterances processed: 2,092
    \item Number of utterances about contemporary events: 479
    \item Number of utterances not about contemporary events: 1,613
    \item Percentage of utterances about contemporary events: 22.90\%
    \item Percentage of utterances not about contemporary events: 77.10\%
\end{itemize}

\subsection{Discussion}
This empirical figure (22.9\%) diverges considerably from earlier claims in the literature suggesting that up to 70\% of conversational content pertains to contemporary events. While lower than the literature estimate, this proportion still represents a substantial share of user utterances, underscoring the relevance of recency in conversational settings. The discrepancy highlights the potential for such estimates to be context-dependent or to overgeneralize across different dialogue scenarios. Future work could strengthen these findings by examining the prevalence of contemporary references across multiple corpora or dialogue domains, thereby assessing their consistency and generalizability.

\section{Prompt given to LLM}
\label{appendix: Prompt given to LLM}

\subsection{Prompt for judging whether the response is correct or not}
\label{appendix: Prompt for judging whether the response is correct or not}

We used the following prompt to judge whether the provided answer by a system is correct or not.

\begin{Verbatim}
Judge whether the answer to the following question is correct 
based on the provided label answer.
Here is the question: {question}
Here is the label answer {label_answer}
Here is the answer that you will be judging: {answer_to_judge}
If the answer to judge is correct, output 1, if it's incorrect, output 0.
If the answer to judge is something like "I don't know" or 
"No information found", output 2.
Output 1 even if the answer is only partly correct.
Output 0 if the answer uses "I" as the subject, 
because the question is about the user,not the assistant.
Only output 1 or 0 or 2, nothing else, no explanation needed.
    
\end{Verbatim}

\subsection{Prompt for Categorizing Utterances as Profile-related, Episode-related, or Other}
\label{appendix: prompt_profile_episode}

We used the following prompt to classify each user utterance into one of three categories: profile-related, episode-related, or other. The model's output was used for post-hoc analysis of utterance types.

\begin{verbatim}
You will be given a single user utterance from a 
human-chatbot conversation.

Classify the utterance into one of the following three categories:

Profile-related: The user is stating biographical 
facts, preferences, personal attributes,
or other stable information about themselves 
(e.g., "My favorite color is blue",
"I have two brothers", "I live in New York").

Episode-related: The user is referring to a specific 
event or experience,
whether in the past, present, or future 
(e.g., "I went to a concert last night",
"I'm going to Hong Kong next month", 
"I had a fight with my roommate").

Other: The utterance does not fall into the above categories. 
This includes small talk,
greetings, questions, or general opinions not 
tied to the user’s identity or experiences.

Please output only the category label: 
"Profile-related", "Episode-related", or "Other".

Here is the utterance:
"{user_utterance}"
\end{verbatim}

\section{Cosine Similarity Threshold}
\label{appendix: Optimal Cosine Similarity Threshold}

We used the default value of 0.8 for the cosine similarity threshold. Using the LUFY-Dataset, we conducted experiments to verify whether this choice was optimal.

We used the LUFY dataset, which consists of 510 QA pairs. Of these, only 2.9\% (15 questions) were multi-hop questions requiring at least two pieces of information. Therefore, we analyzed the remaining 495 single-hop questions—for example, “What’s my favorite movie?”, which can be answered by retrieving a single utterance (e.g., "My favorite movie is Interstellar") from the conversation.

For each pair, we retrieved ten relevant memories with a cosine similarity threshold of at least 0.6 and annotated whether the retrieved memory was correct for answering the question. The correctness of a retrieved answer was largely objective, as it involved verifying whether the retrieved memory semantically matched the ground truth answer. One annotator conducted this verification, and since it was an objective comparison, we did not compute inter-annotator agreement metrics such as Cohen’s kappa. In the rare case of ambiguity, disagreements were resolved through discussion among the authors. 

\begin{figure}[ht!]
\centering
\includegraphics[width=0.8\textwidth]{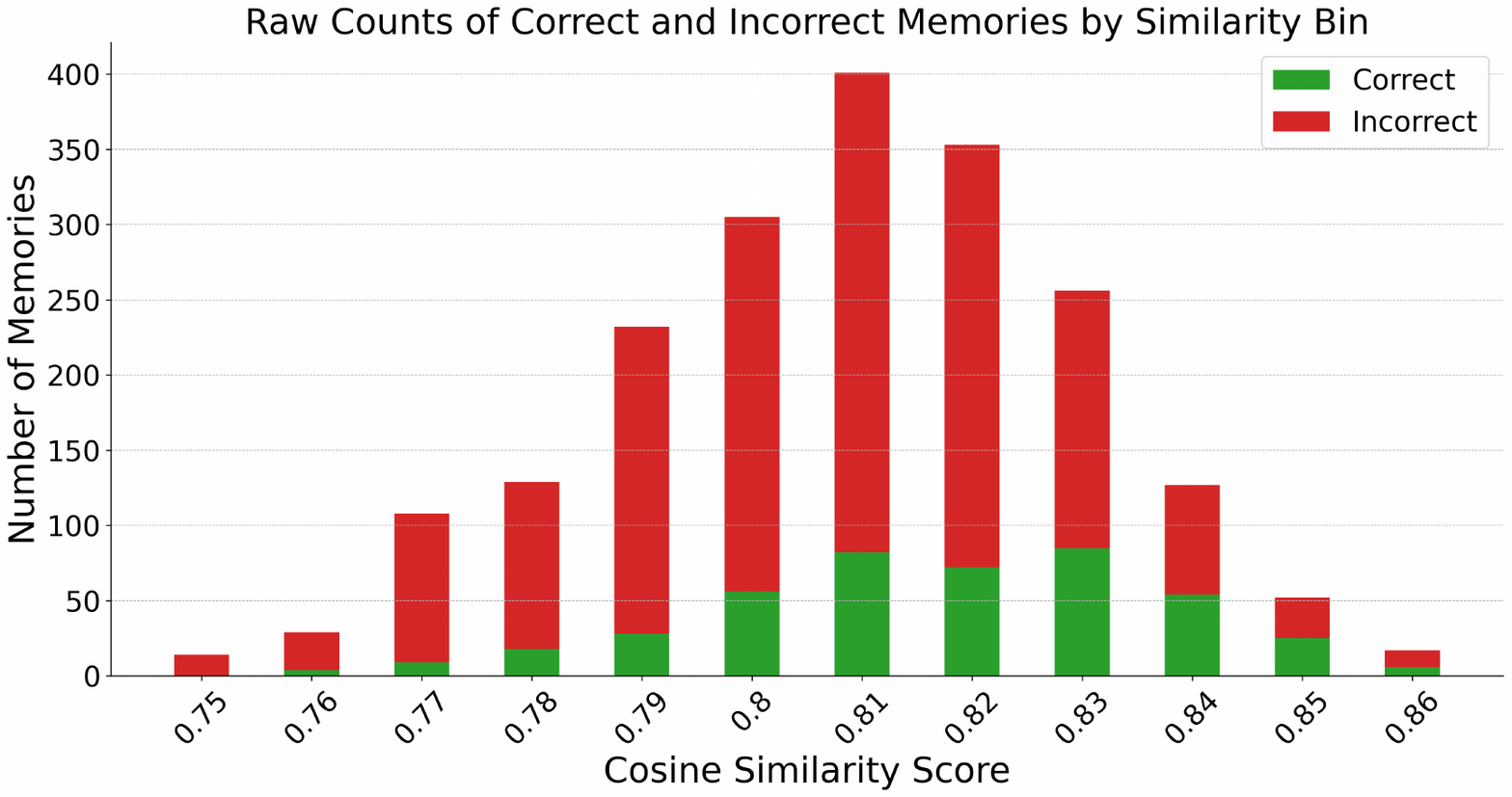}
\caption{Percentage of correct memories for different thresholds. Thresholds below 0.75 are omitted because in our analysis, no correct memories were retrieved in that range.}
\label{fig: Correct_SimScore}
\end{figure}

Figure~\ref{fig: Correct_SimScore} shows the raw counts of correct and incorrect memories across cosine similarity score bins, illustrating how both relevant and irrelevant memories are distributed across similarity ranges. Our findings include:

\begin{itemize}
\item Memories with a cosine similarity of less than 0.75 were always irrelevant or incorrect.
\item There is a positive correlation between higher cosine similarity and the likelihood of the memory correctly answering the question, although fewer cases are retrieved when the cosine similarity is over 0.85.
\end{itemize}

Based on these findings, we tested various cosine similarity thresholds between 0.75 to 0.83 to determine the threshold that would yield the highest F1 Score, representing the balance between retrieving relevant information and filtering out irrelevant content. 

\begin{figure}[ht!]
\centering
\includegraphics[width=0.8\textwidth]{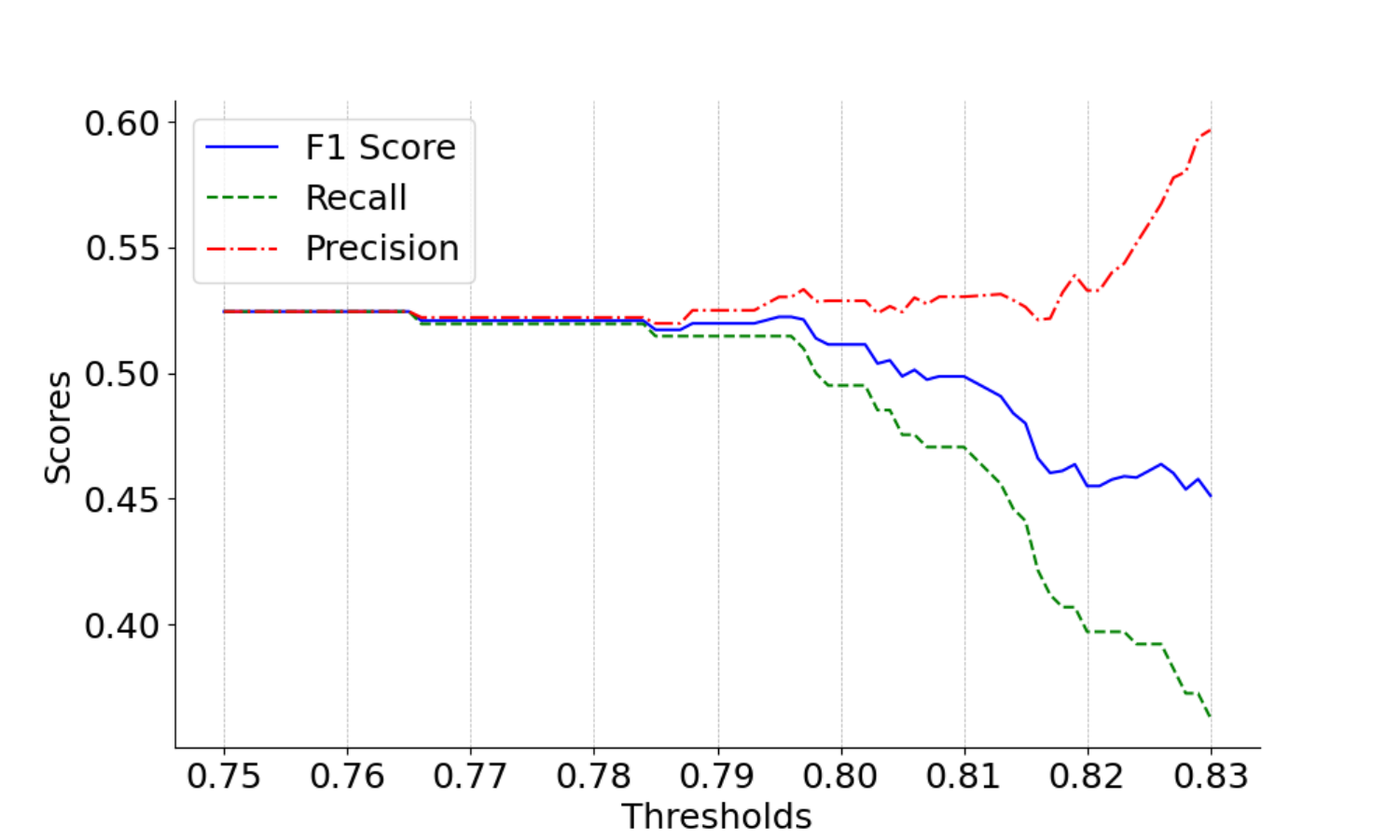}
\caption{F1 Score, Recall, and Precision at various thresholds. Minimal change in F1 Score between 0.75 to 0.795, with a significant decline beyond 0.8.}
\label{fig: F1_Thresholds}
\end{figure}

As shown in Figure~\ref{fig: F1_Thresholds}, we observed that F1 Score remains stable between thresholds of 0.75 and 0.8, with a notable drop beyond 0.8. While F1 Score was used to characterize the overall trade-off between precision and recall, we recognize that the goal of retrieval in this context—answering single-hop memory questions—may prioritize high-precision retrieval over exhaustive recall. Indeed, in most cases, retrieving a single correct memory suffices for generating an accurate response.

Nevertheless, we selected 0.8 as a practical threshold because it retains a strong balance: it filters out many irrelevant memories (as seen with poor performance below 0.75) while still admitting a sufficient number of relevant ones to answer most questions. Moreover, this value aligns with commonly used defaults in RAG frameworks such as LlamaIndex, providing a reasonable and reproducible baseline. We acknowledge that for tasks with stricter precision requirements, a higher threshold might be preferable, and exploring task-specific threshold tuning remains a promising direction for future work.

\clearpage

\section{Instructions given to the annotators for Subjective Evaluations}
\label{appendix: instructions for annotators for subjective eval}

We provide the specific instructions for the three criteria given to annotators.
\subsubsection*{Did the chatbot's personalization appear appropriate in the conversation?}

\begin{enumerate}
    \item \textbf{1/5 (Very Poor):} Responses feel generic and lack personalization.
    \item \textbf{2/5 (Poor):} Limited personalization attempts are often off-target or superficial. Responses only occasionally reflect the user's context.
    \item \textbf{3/5 (Average):} The chatbot shows moderate personalization. Responses are relevant but may not fully address specific user needs.
    \item \textbf{4/5 (Good):} The chatbot consistently personalizes well, matching responses to the user's context and needs effectively.
    \item \textbf{5/5 (Excellent):} Outstanding personalization. Responses are highly relevant, context-aware, and perfectly meet the user's needs, enhancing engagement and satisfaction.
\end{enumerate}

\subsubsection*{How well did the conversation flow without feeling disjointed or out of context?}

\begin{itemize}
    \item \textbf{1/5 (Very Poor):} The conversation feels broken and illogical, with responses often out of context.
    \item \textbf{2/5 (Poor):} The conversation has frequent awkward transitions or non-sequiturs that disrupt the flow.
    \item \textbf{3/5 (Average):} The conversation flows reasonably well, with some disjointed moments that slightly distract from the overall experience.
    \item \textbf{4/5 (Good):} The conversation flows well, with only minor issues that do not significantly impact the user's experience.
    \item \textbf{5/5 (Excellent):} The conversation flows seamlessly and logically, feeling completely natural and coherent throughout.
\end{itemize}

\subsubsection*{Overall, how would you rate the user's experience with the chatbot?}

\begin{enumerate}
    \item \textbf{1/5 (Very Poor):} The user found the conversation frustrating and unhelpful, strongly feeling they would not want to use the chatbot again.
    \item \textbf{2/5 (Poor):} The user was somewhat disappointed with the conversation, finding little value in it and is unlikely to use the chatbot again soon.
    \item \textbf{3/5 (Average):} The conversation met the user's basic expectations. They would consider using the chatbot again if needed.
    \item \textbf{4/5 (Good):} The user was pleased with the conversation and found it helpful, expressing a clear interest in using the chatbot again.
    \item \textbf{5/5 (Excellent):} The user was highly satisfied with the conversation, finding it exceptionally useful and engaging, and is eager to use the chatbot again.
\end{enumerate}

To assess the consistency of human judgments across subjective criteria---\textit{Personalization}, \textit{Flow of Conversation}, and \textit{Overall Quality}---we computed the Intraclass Correlation Coefficient (ICC) across all annotators for each metric. The resulting ICC values were:

\begin{itemize}
    \item \textbf{Personalization}: 0.31
    \item \textbf{Flow of Conversation}: 0.29
    \item \textbf{Overall Quality}: 0.32
\end{itemize}

These values indicate fair to low inter-rater agreement. While relatively low, such results are consistent with previous findings in dialogue evaluation literature, where subjective judgments often yield low reliability due to inherent variability in interpretation and preference.

For example,~\citep{braggaar2022reproduction} highlighted that Likert-scale evaluations in dialogue systems frequently suffer from inconsistency, especially when raters lack a shared rubric or reference standard. Similarly, ICC values below 0.50 were reported in expert evaluations of conversational quality~\citep{gmel2020should}, and other dialogue-related tasks---such as behavior coding~\citep{howell2023examination} and clinical summarization~\citep{fraile2025expert}---have shown similarly low reliability due to ambiguous criteria, narrow rating variance, and inconsistent calibration.

\section{Complete Results of the Evaluations with the QA Pairs}
\label{appendix: Complete Results}

\clearpage
\begin{table}
    \centering
    \begin{tabular}{lccccccccccccc}
    \hline
     & \multicolumn{1}{c}{S1} & \multicolumn{1}{c}{} & \multicolumn{2}{c}{S2} & \multicolumn{1}{c}{} & \multicolumn{3}{c}{S3} & \multicolumn{1}{c}{} & \multicolumn{4}{c}{S4} \\ 
     System & Q1 & & Q1 & Q2 &  & Q1 & Q2 & Q3 & & Q1 & Q2 & Q3 & Q4 \\ \hline
    Naive RAG & 75.6 & & 68.2 & 71.0 & & 64.3 & 61.3 & 97.1 & & 62.5 & 62.1 & \textbf{91.7} & 63.3 \\
    MemoryBank & 63.6 & & 56.7 & 44.8 & & 60.0 & 40.9 & 71.4 & & 63.3 & 43.5 & 63.6 & 88.0 \\
    LUFY & \textbf{80.0} & & \textbf{81.3} & \textbf{92.6} & & \textbf{82.8} & \textbf{77.8} & \textbf{100.0} & & \textbf{82.1} & \textbf{88.9}  & 83.3 & \textbf{92.0} \\ \hline
        \end{tabular}     \vspace{10pt}
    \caption{Complete results showing the Precision scores of each system across different sessions and question sets. Q1 denotes Questions about session1, Q2 denotes Questions about session2 and so on.}
    \label{tab:Full Precision Table}
\end{table}

\begin{table}
    \centering
    \begin{tabular}{lccccccccccccc}
    \hline
    & \multicolumn{1}{c}{S1} & \multicolumn{1}{c}{} & \multicolumn{2}{c}{S2} & \multicolumn{1}{c}{} & \multicolumn{3}{c}{S3} & \multicolumn{1}{c}{} & \multicolumn{4}{c}{S4} \\ 
     System & Q1 & & Q1 & Q2 &  & Q1 & Q2 & Q3 & & Q1 & Q2 & Q3 & Q4 \\ \hline
    Naive RAG & \textbf{60.8} & & \textbf{58.8} & 43.1 & & \textbf{52.9} & \textbf{37.3} & \textbf{64.7} & & \textbf{49.0} & \textbf{35.3} & \textbf{64.7} & 37.3 \\
    MemoryBank & 41.2 & & 33.3 & 25.5 & & 35.3 & 17.6 & 39.2 & & 37.3 & 19.6 & 27.5 & 43.1 \\
    LUFY & 54.9 & & 53.0 & \textbf{49.0} & & 47.1 & 27.5 & 41.2 & & 45.1 & 31.4 & 19.6 & \textbf{45.1} \\ \hline
        \end{tabular}     \vspace{10pt}
    \caption{Complete results showing the Recall scores of each system across different sessions and question sets.}
    \label{tab:Full Recall Table}
\end{table}

\begin{table}
    \centering
    \begin{tabular}{lccccccccccccc}
    \hline
     & \multicolumn{1}{c}{S1} & \multicolumn{1}{c}{} & \multicolumn{2}{c}{S2} & \multicolumn{1}{c}{} & \multicolumn{3}{c}{S3} & \multicolumn{1}{c}{} & \multicolumn{4}{c}{S4} \\ 
     System & Q1 & & Q1 & Q2 &  & Q1 & Q2 & Q3 & & Q1 & Q2 & Q3 & Q4 \\ \hline
    Naive RAG & \textbf{67.4} & & 63.2 & 53.6 & & 58.0 & \textbf{46.4} & \textbf{77.7} & & 54.9 & 45.0 & \textbf{75.9} & 46.9 \\
    MemoryBank & 50.0 & & 42.0 & 32.5 & & 44.4 & 24.6 & 50.6 & & 46.9 & 27.0 & 38.4 & 57.9 \\
    LUFY & 65.1 & & \textbf{64.2} & \textbf{64.1} & & \textbf{60.0} & 40.6 & 58.4 & & \textbf{58.2} & \textbf{46.4} & 31.7 & \textbf{60.5} \\ \hline
        \end{tabular}     \vspace{10pt}
    \caption{Complete results showing the F1 scores of each system across different sessions and question sets.}
    \label{tab:Full F1 Table}
\end{table}

\clearpage
\section{Illustrative Example of Memory Retention and Retrieval Across Systems}
\label{appendix: Illustrative Example of Memory Retention and Retrieval Across Systems}

To illustrate the characteristic behavior of the three systems, we present a constructed scenario highlighting how each handles memory retention and retrieval in response to a user’s emotionally significant utterance. The key memory—“I got into Harvard”—was originally mentioned in Session 2.

\begin{table}[ht]
\centering
\renewcommand{\arraystretch}{1.3}
\setlength{\tabcolsep}{8pt}
\begin{tabular}{|p{14.5cm}|}
\hline
\textbf{System: Naive RAG} \\
\textit{Memory status:} Retained (no forgetting). \\
\textit{Retrieval:} Fails to retrieve the "Harvard" memory due to low lexical overlap with the user’s input. \\
\textit{Bot Response:} “That’s understandable. College can be stressful. Want to talk more about it?” \\
\hline

\textbf{System: MemoryBank} \\
\textit{Memory status:} Forgotten. Not retrieved in Session 2, so decayed due to low frequency and recency. \\
\textit{Retrieval:} Memory deleted in forgetting phase. \\
\textit{Bot Response:} “Yeah, college planning can be tough. Do you know where you’re applying yet?” \\
\hline

\textbf{System: LUFY (Ours)} \\
\textit{Memory status:} Retained due to high emotional arousal score. \\
\textit{Retrieval:} Importance-boosted retrieval surfaces the Harvard memory despite low lexical similarity. \\
\textit{Bot Response:} “Got into Harvard—no wonder you're feeling pressure. Want to talk?” \\
\hline
\end{tabular}
\vspace{10pt}
\caption{
Comparison of system responses in Session 3 after the user says: 
\textit{“I’m feeling overwhelmed lately. So much pressure with upcoming college stuff.”} 
Only LUFY successfully recalls the emotionally important but unrepeated “I got into Harvard” memory, which was originally mentioned in Session 2.
}
\label{tab:harvard_example_case_analysis}
\end{table}

\end{document}